\documentclass[11pt]{article}

\usepackage[final]{acl}

\usepackage{times}
\usepackage{latexsym}
\usepackage[T1]{fontenc}
\usepackage[utf8]{inputenc}
\usepackage{microtype}
\usepackage{inconsolata}
\usepackage{graphicx}
\usepackage{booktabs}
\usepackage{amsmath}
\usepackage{amssymb}
\usepackage{multirow}
\usepackage{makecell}
\usepackage{xcolor}

\title{How You Ask Shapes What You Get: A Theory-Seeded Measurement of Articulation in Advice-Seeking LLM Conversations}

\author{Juneha Baek, Suhyeon Lee, Donghyuk Shin\thanks{\ Corresponding author.} \\
  KAIST \\
  \texttt{\{juneha.baek, suhyeonlee, dhs\}@kaist.ac.kr} \\}

\begin{document}
\maketitle

\begin{abstract}
Users articulate the same advice-seeking request in different ways: some specify detailed constraints, others gesture at a vague need. Prior work treats this variation as noise to be averaged away; we instead treat it as a stable, measurable structure in the input distribution. We ask whether articulation (\emph{how} people ask) forms latent dimensions separable from topic (\emph{what} they ask about), and whether it is associated with how language models respond. We extract interpretable features from $16{,}447$ advice-seeking prompts pooled from public chat corpora (WildChat, LMSYS, and ShareChat) and recover a small set of latent articulation factors that replicate across train/test splits and across corpora. Because this structure is largely separable from topic, the populations it defines cut across topics and stay invisible to topic- or task-based evaluation. The factors define a handful of recurring articulation styles, one of which stands out: a long-form but information-poor style, roughly one in six prompts in the largest corpus, where models return shorter, vaguer answers and do not ask for clarification even though under-specification is exactly the condition that warrants it. The contrast holds within every topic group and length quintile, and is not under-specification alone --- a second, equally under-specified style \emph{does} draw clarifying questions. Two independent human annotators reproduce this contrast. We argue that benchmarks should stratify on articulation, and we offer the extracted structure as a measurement instrument for doing so.
\end{abstract}

\section{Introduction}
\label{sec:intro}

The same advice request can be expressed in radically different ways. One user asks ``\emph{recommend a quiet under-\$1500 laptop with at least 16GB RAM and 14-inch display for academic writing}''; another asks ``\emph{any laptop you'd suggest? not really sure what I need}''. Both seek a laptop recommendation; their articulations differ on dimensions that are not about the topic: how specific, how hard the constraints, how many named alternatives, how hedged.

Existing work treats this variation either as noise to be removed (paraphrase robustness, \citealp{sclar2024sensitivity, mizrahi2024state, webson2022prompt, lu2022fantastically}) or as a problem of LLM behavior (\citet{shaikh2024grounding} report that LLMs perform far fewer grounding acts, such as clarification and acknowledgment, than humans do in comparable conversations). Both framings are downstream of a more basic question that, to our knowledge, has not been answered for naturalistic chat: does articulation variation in the wild form a stable structure that is separable from topic? The answer determines what an LLM evaluation can even see. If articulation collapses onto topic, then varying prompts across topics already covers it and nothing is missed; if instead it is a separate, low-dimensional axis, then a whole dimension of how people ask, and of how models respond, runs orthogonal to the topic and task axes that benchmarks are built on, and is silently absent from them. We therefore also ask how many latent dimensions the structure spans, and whether it systematically covaries with what LLMs return.

Our central move is to treat articulation not as noise to be averaged away but as a stable structure to be measured. We answer the questions above descriptively on $16{,}447$ advice-seeking first-turn prompts pooled from public chat corpora (WildChat, LMSYS, and ShareChat),\footnote{Code, configs, and tables: \url{https://github.com/Juneha-Baek/articulation-segmentation-emnlp2026}.} and find that articulation has a stable, low-dimensional structure that replicates across train/test splits and, for the core dimensions, across corpora, and is largely orthogonal to topic: a genuinely separate axis of prompt variation that topic- and task-based analysis does not capture.

Why the axis matters comes first, the instrument second. There is a population of advice prompts, roughly one in six, that is long but information-poor: verbose context wrapped around a diffuse ask. For these, models return shorter, vaguer answers with fewer concrete recommendations and, tellingly, no clarifying question (Section~\ref{sec:impl}) --- the input-side, population-scale counterpart to the clarification deficit \citet{shaikh2024grounding} document on hand-curated prompts. This population is not a topic, not a task, and not a length band; it cuts across all three, so evaluation organized by topic or task never sees it, and a model that fails on under-articulated prompts scores better than it should. Making it visible requires an instrument, and that is what we build. Concretely, we (i) show that articulation variation in naturalistic chat has a stable, low-dimensional latent structure that replicates across corpora (Section~\ref{sec:r1}-\ref{sec:r2}); (ii) show that this structure is largely separable from topic (Section~\ref{sec:r3}); and (iii) use it to locate and characterise that population (Section~\ref{sec:impl}). The implication is that benchmark evaluation should stratify on articulation, not topic alone.

Under-specification can of course be scored per prompt, and one feature detects our at-risk prompts at $\text{AUC}{=}0.91$. But the claims above are distributional and no per-prompt score delivers them: how many axes the variation spans, whether they reproduce out of sample, and what fraction of real traffic is at risk --- the last being what any benchmark-coverage argument needs. Population framing is also what makes the claim falsifiable. This is, finally, a phenomenon-discovery rather than a methods paper: its components are standard and deliberately so, since auditable machinery minimises researcher degrees of freedom.

On scope: chat logs identify the encoded prompt and the model's reply, never the user's unobserved preference $\theta$, so we treat articulation as a measurable property of \emph{prompts} and the link to responses as an association rather than a counterfactual (see Limitations).

\section{Related Work}
\label{sec:related}

\paragraph{Prompt sensitivity and paraphrase robustness.}
\citet{sclar2024sensitivity}, \citet{mizrahi2024state}, and \citet{lu2022fantastically} document that small surface perturbations shift LLM outputs substantially, and \citet{webson2022prompt} show prompt-based models succeed equally on instructive and intentionally misleading prompts, implying surface form has effects not tied to semantic content. This line frames paraphrase robustness as model instability; we instead treat naturalistic articulation variation as a \emph{stable structure} in the input distribution and ask whether it has measurable dimensionality.

\paragraph{Conversational grounding.}
The theoretical basis for clarification traces to \citet{clark1989contributing}'s grounding framework, in which speakers collaboratively establish mutual understanding through clarification and repair. \citet{shaikh2024grounding} show LLMs generate far less grounding than humans, and \citet{shaikh2025grounding} quantify the gap on naturalistic data ($16\times$ less likely to provide follow-up requests). Methods for asking effective clarifying questions are an established line \citep{rao2018learning, aliannejadi2019asking, zamani2020generating}; we identify the input-side articulation structure this deficit responds to, and show the population most warranting clarification is identifiable at the input level.

\paragraph{Ambiguity, underspecification, and preference elicitation.}
A parallel line studies ambiguous or underspecified prompts: \citet{min2020ambigqa} find over half of open-domain questions ambiguous, benchmarks test whether LLMs identify the minimal missing information rather than answering prematurely \citep{li2025questbench}, \citet{li2023eliciting} use LMs to elicit ``nebulous'' preferences through free-form interaction, and \citet{sarkar2025prompt} show on real human-LLM logs that user queries frequently fall short of the underlying information need. This work treats underspecification largely per-prompt (is \emph{this} prompt answerable?); we ask whether under-articulation is a \emph{population-level} structure (a reproducible segment of the input distribution) and measure how responses covary with it at scale.

\paragraph{User segmentation in LLM use.}
Most prior segmentation of LLM users is by \emph{topic} or \emph{task}: the ShareChat taxonomy \citep{yan2025sharechat}, the WildChat domain analysis \citep{zhao2024wildchat}, the task-oriented analysis of \citet{ouyang2024shifted}, the consumer-usage analysis of \citet{chatterji2025chatgpt}, and Clio \citep{tamkin2024clio}. \citet{chatterji2025chatgpt} find advice-seeking (``Asking'') is roughly half of consumer messages, underscoring the scale of the population we study. We instead segment on articulation (\emph{how} users phrase their asks) and show the two questions yield largely separable partitions.

\paragraph{Consumer-preference theory as feature inspiration.}
We borrow Nelson's search/experience trichotomy \citep{nelson1970information} (extended to credence attributes by \citealp{darby1973credence}) for attribute type, Bettman's constructed-preference view \citep{bettman1998constructive, payne1993adaptive} for crystallization (whether preferences look settled or still forming), and Shannon's channel notion \citep{shannon1948mathematical} for articulation bandwidth. These are feature sources, not constructs to be validated: the latent structure is determined by factor analysis, not by theory.

\section{Method}
\label{sec:method}

\paragraph{Why not simply ask an LLM?}
One could ask a current LLM, per prompt, how well articulated it is. We instead use the LLM only for local span extraction (Section~\ref{sec:s1}) and leave discovery to unsupervised statistics, because a zero-shot judgment presupposes the label it applies: it can recover a category we already know to name, but not tell us how many dimensions the variation spans, whether they reproduce across corpora, or surface a configuration nobody named in advance.

\subsection{Corpora and Filtering}
\label{sec:s0}

\textbf{What counts as an advice-seeking prompt.} We study first-turn messages in which the user asks the assistant to \emph{help them decide something about their own situation}: to pick between options, to recommend a product, service, place, plan, or course of action, to advise on a personal situation, or to delegate such a decision --- usually, though not necessarily, with some preference, constraint, or context stated. This excludes task instructions (write, translate, summarise, roleplay), factual or how-to questions with no personal stake, and persona setup. This definition, not a keyword list, is what the Stage~0 classifier operationalises (verbatim prompt in Appendix~\ref{sec:appendix-prompts}, E.1).

\textbf{Corpora and their roles.} We pool first-turn user prompts (plus the immediate assistant response) from three public chat corpora --- WildChat-4.8M \citep{zhao2024wildchat}, LMSYS-Chat-1M \citep{zheng2024lmsys}, and ShareChat \citep{yan2025sharechat} --- restricting to English-only conversations after each corpus's language filter. ShareChat is treated as one pooled corpus throughout; it ships as five platform logs, and Tables~\ref{tab:s0-prevalence} and~\ref{tab:replication} report that breakdown as a transparency note and as a way of varying the responding assistant while holding corpus construction fixed. The three roles are deliberately asymmetric (Table~\ref{tab:s0-prevalence}): testing generalisation requires the structure be \emph{learned} on one corpus and only \emph{applied} to the others, so WildChat is the sole training corpus and nothing downstream of Stage~3 is ever fit on LMSYS or ShareChat.

Not every chat-corpus prompt is an advice-seeking request. To recover the subset that is, we use a two-pass cascade.

\textbf{1st pass: domain net.}
A high-recall keyword cascade with seven domain matchers (shopping, travel, career, finance, content/media, lifestyle/food, plus a domain-agnostic advice-seeking pseudo-domain) admits any prompt with a domain-relevant trigger. To recover advice prompts that escape lexical match, we additionally run \texttt{intfloat/e5-large-v2} \citep{wang2022e5} against $23$ advice-seeking anchor prototypes and admit any prompt with maximum cosine similarity $\geq 0.80$. The threshold is calibrated on a held-out sample because e5 has non-trivial baseline similarity ($p_{50}{\approx}0.77$).

\textbf{2nd pass: intent LLM.}
A task-pattern pre-filter (regex matching transformation imperatives, persona setup, prompt injection, fiction/roleplay openers, etc.) drops obvious task prompts before LLM classification. Surviving candidates are classified by \texttt{gpt-4o-mini} as \texttt{is\_recommendation} $\in \{0,1\}$ under a prompt requiring the input to be an advice/recommendation/delegation under stated preferences.

Out of $1{,}768{,}604$ loaded English first-turn pairs, the cascade admits $535{,}612$ candidates ($30.3\%$); the LLM 2nd pass keeps $16{,}447$ positives (Table~\ref{tab:s0-prevalence}). False-negative rate on a stratified sample of cascade rejects is $0.4$--$1.6\%$ across corpora, suggesting the cascade does not systematically miss advice prompts. Domain keywords, the $23$ e5 advice anchors, and the threshold calibration are in Appendix~\ref{sec:appendix-stage0}.

\begin{table}
\centering\footnotesize
\setlength{\tabcolsep}{3pt}
\begin{tabular}{lrrrr}
\toprule
Dataset & Loaded & Cand. & LLM+ & Prev. \\
\midrule
WildChat        & 1{,}211{,}454 & 414{,}406 & \textbf{7{,}225} & 0.60\% \\
LMSYS           &   479{,}413 & 104{,}547 & \textbf{7{,}720} & 1.61\% \\
SC-ChatGPT      &    56{,}078 &  12{,}907 & 1{,}163 & 2.07\% \\
SC-Grok         &     8{,}594 &   1{,}758 &    165 & 1.92\% \\
SC-Perplexity   &     9{,}410 &   1{,}236 &    111 & 1.18\% \\
SC-Gemini       &     2{,}990 &      603 &     51 & 1.71\% \\
SC-Claude       &        665 &      155 &     12 & 1.80\% \\
\midrule
Total           & 1{,}768{,}604 & 535{,}612 & \textbf{16{,}447} & 0.93\% \\
\bottomrule
\end{tabular}
\caption{Stage~0 prevalence and the role of each slice. \emph{SC} = ShareChat; \emph{Loaded}\ = English first-turn prompts retained after de-duplication and a minimum-length filter; \emph{Cand.}\ = cascade candidates; \emph{LLM+}\ = surviving the LLM intent classifier. WildChat is the sole \textbf{training} corpus: the factor model and the GMM are fit there and only there. LMSYS is the \textbf{primary replication} corpus, held out and receiving the fitted model transform-only, and cross-corpus congruence (Table~\ref{tab:factors}) is measured against it. The five SC slices, pooled, give \textbf{cross-assistant replication} under the same transform-only application; SC-Claude ($n{=}12$) is too small for segment replication and appears here only.}
\label{tab:s0-prevalence}
\end{table}

\subsection{Articulation Profile}
\label{sec:s1}

For each positive prompt $p$, we extract an articulation profile $a(p)$ via LLM span tagging (\texttt{claude-haiku-4-5}, temperature $0$, JSON-structured output). The LLM is prompted to enumerate intent spans, label each span with attribute type (search/experience/credence; \citealp{nelson1970information, darby1973credence}), hardness (hard/soft/exemplar/open; \citealp{bettman1998constructive}), specificity ($1$--$5$), and brand/exemplar flags, and to provide prompt-level features (crystallization $1$--$5$, hedge markers, sentence mood, person distribution, imperative count, named alternatives). Span counts and per-prompt features yield a flat numeric vector of $30+$ raw features per prompt. After Stage~3 preprocessing (Section~\ref{sec:s3}), $19$ features enter factor analysis; the full feature list and the preprocessing pipeline are in Appendix~\ref{sec:appendix-features}.

We extract features as a \emph{candidate set} rather than as a pre-committed set of dimensions. The latent dimensionality is determined by factor analysis, not by us. The full LLM prompt templates are in Appendix~\ref{sec:appendix-prompts}.

\subsection{Response Profile}
\label{sec:s2}

For each prompt $p$ we also measure an outcome vector $r(p)$ on the assistant's reply, extracted by the same LLM: number of recommended items, clarifying questions, attribute coverage (whether search/experience/credence attributes are addressed), recommendation specificity, hedge markers, whether a direct answer was given, and recommendation diversity. These outcomes are the dependent variables in the response analysis (Section~\ref{sec:impl}).

\subsection{Factor Discovery}
\label{sec:s3}

We preprocess the raw feature matrix by applying isometric log-ratio (ilr) transform \citep{egozcue2003isometric} to the compositional groups (attribute ratios that sum to $1$, hardness ratios) to break the simplex constraint that compositional data otherwise impose \citep{aitchison1986statistical}, length-normalizing count features by token count, dropping near-zero-variance columns, and standardizing. The final preprocessed matrix has $19$ features.

\textbf{Choosing $k$.} The number of factors is set by parallel analysis \citep{horn1965parallel} rather than by us, and it comes out at $k{=}6$ (Figure~\ref{fig:scree}). The procedure, the two random benchmarks we check it against, and the eigenvalues at the $k{=}6/k{=}7$ boundary are in Appendix~\ref{sec:appendix-modelsel}.

\textbf{Factor extraction and rotation.} We fit a $6$-factor model with principal axis factoring and Promax (oblique) rotation, allowing inter-factor correlation since articulation dimensions need not be uncorrelated.

\subsection{Segmentation}
\label{sec:s4}

On the WildChat training set ($n{=}7{,}225$) projected into the $6$-factor space, we fit a Gaussian Mixture Model with the number of components capped at the factor count. BIC selects $k{=}6$ articulation modes. The full $k$-selection curves, the reason for the cap, and the reason we read BIC rather than silhouette in this rotated space are in Appendix~\ref{sec:appendix-modelsel}.

\textbf{Replication.} The WildChat-trained GMM is then \emph{applied} to the held-out non-WildChat corpora in transform-only mode, so the segment definitions are fixed and only the empirical segment distribution varies across corpora.

\subsection{Topic Re-tagging}
\label{sec:s5}

To put all three corpora on a comparable topic basis (ShareChat ships with platform-specific labels not directly comparable to WildChat or LMSYS), we re-tag every prompt with a unified topic taxonomy adapted from \citet{yan2025sharechat}: $25$ fine-grained topics rolling up to $7$ high-level groups (H1 Information \& News, H2 Personal Decisions \& Recommendations, H3 Education, H4 Technical/STEM, H5 Creative \& Media, H6 Relational \& Emotional, H7 Other \& General; full mapping in Appendix~\ref{sec:appendix-topics}). The original ShareChat labels are not reused for analysis.

\subsection{From Profiles to Claims}
\label{sec:claims}

Each prompt $p$ now carries an articulation profile $a(p)$, an outcome vector $r(p)$ measured on the reply, a topic label $z(p) \in \{\mathrm{H1},\dots,\mathrm{H7}\}$, and a segment $s(p)$, the most probable of the $K{=}6$ GMM components given $a(p)$. Our three questions become three concrete tests. \textbf{Structure}: does articulation cluster into recurring types at all, and does the partition survive resampling (bootstrap ARI, Section~\ref{sec:r2})? The claim is about the \emph{partition}, not the individual, since the corpora do not track users. \textbf{Separability}: is $s$ distinct from $z$? Cramér's $V$ summarises the dependence, but the decisive test is whether re-clustering \emph{within} each topic recovers the pooled partition (Section~\ref{sec:r3}). \textbf{Covariation}: does segment membership predict the reply once topic, assistant model, corpus, and prompt length are controlled (Section~\ref{sec:impl})? Because $n$ is large we read that through effect sizes rather than $p$-values, and report it as an association: a causal claim would require holding the user's latent goal $\theta$ fixed, which chat logs do not permit. Appendix~\ref{sec:appendix-estimands} states the three estimands formally.

\section{Results}
\label{sec:results}

Each finding is stated in plain terms first, with statistical justification kept to marked paragraphs and the appendix. The map from claim to evidence is: articulation spans six latent dimensions that reproduce out of sample and across corpora (\S\ref{sec:r1}; Appendices~\ref{sec:appendix-modelsel},~\ref{sec:appendix-rerun}), five of whose six names survive a test that could have failed them (Appendix~\ref{sec:appendix-construct}); it recurs as six reproducible but soft-edged segments that are largely separable from topic (\S\ref{sec:r2}--\ref{sec:r3}); and segment membership covaries with response characteristics (\S\ref{sec:impl}; Appendix~\ref{sec:appendix-nested}), sharply so for $s_4$, which gets neither an answer nor a clarifying question and is not simply an ambiguous prompt (Appendix~\ref{sec:appendix-robustness}).

\subsection{Factor stability and cross-corpus replication}
\label{sec:r1}

Each factor has a small number of high-loading features, with top loadings $\lambda$ of $+0.72$ to $+1.02$: F1 specificity/structure (\texttt{ilr\_hard\_2}), F2 articulation density (\texttt{rate\_n\_spans}), F3 named-alternative reference (\texttt{rate\_named\_alt\_count}), F4 search-attribute hardness (\texttt{ilr\_hard\_1}), F5 self-situation disclosure (\texttt{rate\_n\_self\_situation}), and F6 length vs.\ imperative (\texttt{token\_count}). The full $19\times6$ pattern matrix with bootstrap CIs is in Appendix~\ref{sec:appendix-loadings}.

\textbf{How the factors got their names.} The names are ours, assigned after the fact by a mechanical rule we state so it can be checked: for each factor we read off every feature with $|\lambda|>0.30$ in the rotated pattern matrix (Table~\ref{tab:full-loadings}) and named the factor after the feature set dominating it. F2 is \emph{density} because \texttt{rate\_n\_spans} and \texttt{spans\_per\_token} are its only large positive loadings. These name columns of a loading matrix, not validated constructs, and nothing downstream depends on them, since segmentation and every regression operate on factor scores. Labels can still be wrong, so we tested rather than asserted them: a pre-registered known-groups experiment (Appendix~\ref{sec:appendix-construct}) holds on $20$ of $24$ frozen criteria, validating five of six labels at $d{=}1.0$ to $4.6$, at least $9\times$ the extractor's test--retest noise, and falsifying the sixth.

One reading we initially gave F1 does not survive inspection, and we correct it because it was our one claimed confirmation of a seed theory. The extractor's rubric codes crystallization as $1{=}$settled to $5{=}$exploratory, so \emph{low} crystallization means settled. Read correctly, specificity and settledness travel together ($\rho{=}-0.364$): users stating detailed constraints already have formed preferences. That is unremarkable and is not evidence for Bettman's constructed-preference account, so we report it below as a seed hypothesis that did not land.

Internal stability is high for the four core factors (train/test $\phi \geq 0.98$; per-factor values in Table~\ref{tab:factors}, Appendix~\ref{sec:appendix-loadings}), and cross-corpus stability is excellent for the same four (WildChat\,$\leftrightarrow$\,LMSYS $\phi \geq 0.97$). The remaining two, F5 (self-situation) and F6 (length-imperative), sit at $\phi \approx 0.90$ train/test and $\approx 0.82$ cross-corpus, in the ``fair'' band. The imperative/declarative balance and the self-disclosure rate are both corpus-dependent (LMSYS prompts skew shorter and less self-situated), so these two are partly dataset-specific; we retain them because both load heavily on distinct segments downstream.

Two caveats on reading these numbers. Tucker's $\phi$ is a cosine and the textbook bands are calibrated against nothing in particular, so we also report an empirical null: the identical pipeline on column-permuted data returns mean $\phi{=}0.38$ ($95$th percentile $0.67$; Appendix~\ref{sec:appendix-rerun}), which the observed values clear by a wide margin. And a stable measurement bias would recur as cleanly as a stable signal, so $\phi$ cannot detect a reliable-but-invalid extractor (Section~\ref{sec:validation}, Limitations).

\paragraph{Mapping the 6 factors back onto seeding theory.}
The seed theories map onto the factors only partially. Nelson's search attributes form their own factor (F4) while experience/credence markers distribute across F1/F3; Bettman's hardness aligns with F1; and Shannon's bandwidth splits into compact density (F2) and long monologue (F6). Three expectations fail: hedging does not factor out as an independent axis; F5 (self-situation) has no antecedent in our seed; and Bettman's crystallization yields no independent dimension at all, being collinear with specificity. The data thus preserves some theoretical contrasts, collapses others, surfaces a dimension the seed did not predict, and refuses one it did --- failures we report because a theory-seeded design is worth little without them.

\subsection{Articulation segments}
\label{sec:r2}

BIC selects $k{=}6$ articulation density modes. Bootstrap adjusted Rand index (ARI; \citealp{hubert1985comparing}) over $B{=}100$ bootstrap resamples (drawn with replacement) of the training set is $\overline{\mathrm{ARI}}{=}0.549$, $\mathrm{median}{=}0.471$, with $95\%$ CI $[0.377, 0.926]$: the partition is reproducible at the moderate level \citep{steinley2004ari} but with wide variance across resamples, consistent with soft cluster boundaries in a continuous factor space. This figure is conservative: re-running Stage~1 with a stricter extractor and re-executing the pipeline end to end raises it to $0.785$ (Appendix~\ref{sec:appendix-rerun}), so the softness is partly a property of the measurement rather than of the phenomenon.

The six segments are characterized by their factor profiles (examples in Appendix~\ref{sec:appendix-examples}, numeric centroids in Appendix~\ref{sec:appendix-centroids}). \textbf{$s_0$ baseline} ($16\%$) is mid-range on most factors, a generic pattern. \textbf{$s_1$ open-asker} ($15\%$) is very low on F4 (search-hard) with moderate self-situation: open-ended advice not pinned to specifications. \textbf{$s_2$ specifier} ($26\%$, the largest) is high on F1, articulating detailed constraints. \textbf{$s_3$ named-comparer} ($14\%$) is jointly high on F2, F3, and F4 --- dense, brand-aware, spec-driven. \textbf{$s_4$ long-form / low-density} ($16\%$) is very low on F2 (sparse intent spans) and very high on F6 (long declarative monologue): the verbose-but-information-poor population that drives the response-side findings. \textbf{$s_5$ vague + situated} ($14\%$) is very low on F1 with moderate F5, and has the highest clarification rate of any segment.

\begin{figure}[t]
\centering
\includegraphics[width=\columnwidth]{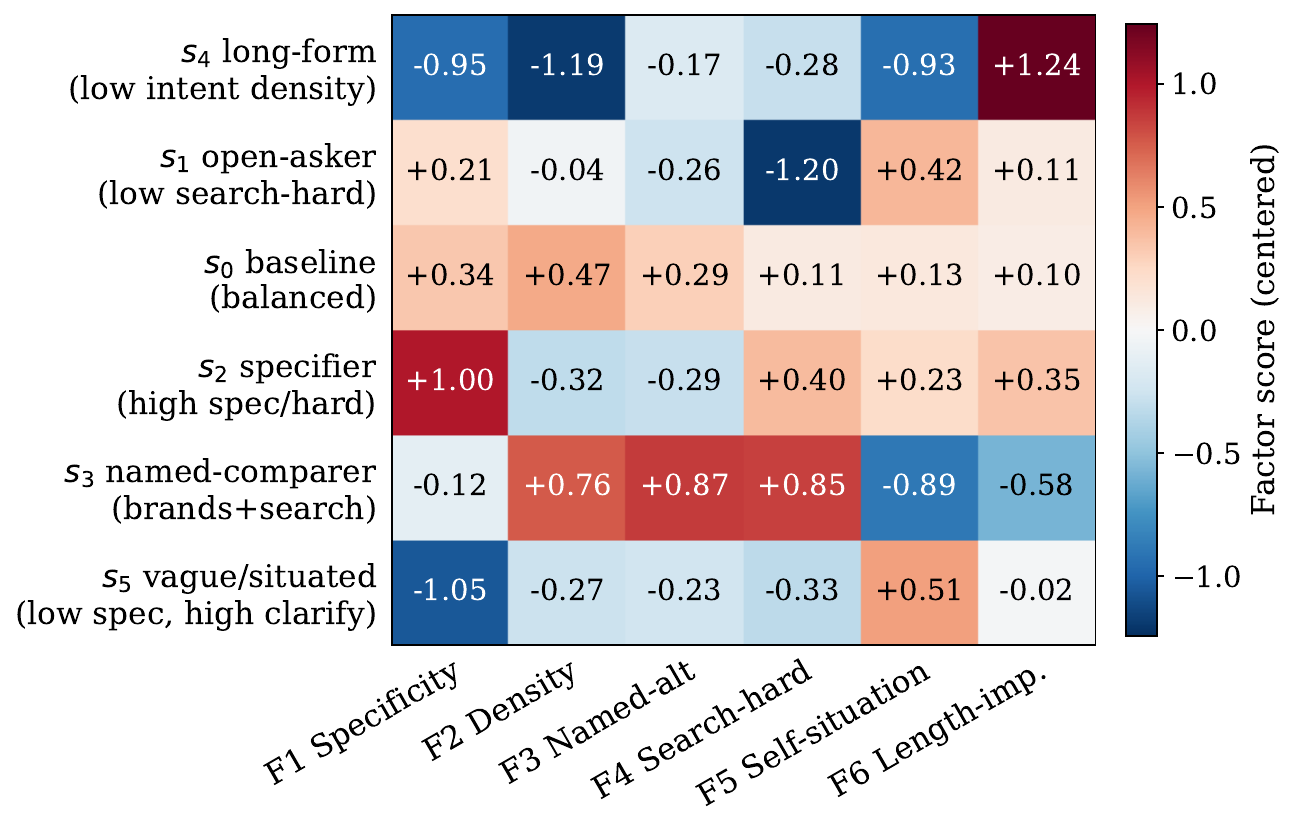}
\caption{Segment centroids on the WildChat training set ($n{=}7{,}225$): $6$ segments $\times$ $6$ factors, each cell the segment's mean standardised factor score, centered at the corpus mean.}
\label{fig:heatmap}
\end{figure}

Figure~\ref{fig:heatmap} shows the centroids in factor space, making each segment's distinguishing factors readable at a glance.

\begin{table*}
\centering\small
\begin{tabular}{lrrrrrrr}
\toprule
Corpus & $n$ & $s_0$ & $s_1$ & $s_2$ & $s_3$ & $s_4$ & $s_5$ \\
\midrule
WildChat (train)          & 7{,}225 & 16\% & 15\% & 26\% & 14\% & 16\% & 14\% \\
LMSYS                     & 7{,}720 & 13\% & 14\% & 21\% & 23\% & \phantom{0}8\% & 21\% \\
ShareChat (ChatGPT)       & 1{,}163 & 15\% & 17\% & 34\% & 12\% & \phantom{0}7\% & 15\% \\
ShareChat (Gemini)        &    51   & 18\% & 16\% & 33\% & 20\% & \phantom{0}2\% & 12\% \\
ShareChat (Grok)          &   165   & 24\% & 13\% & 23\% & 24\% & \phantom{0}4\% & 12\% \\
ShareChat (Perplexity)    &   111   & 20\% & 15\% & 31\% & 21\% & \phantom{0}3\% & 11\% \\
\bottomrule
\end{tabular}
\caption{Segment shares under transform-only prediction with the WildChat-trained GMM. ShareChat is analysed pooled throughout; the four platform rows appear here as a transparency note and to vary the responding assistant with corpus construction held fixed. ShareChat-Claude ($n{=}12$) is below the minimum reliable sample. Row sums can deviate by $\pm 1$\,pp from independent rounding.}
\label{tab:replication}
\end{table*}

Segment shares replicate broadly across corpora (Table~\ref{tab:replication}): $s_2$ is modal everywhere except ShareChat-Grok, and WildChat has the highest $s_4$ share ($16\%$) and the most balanced distribution. The curated ShareChat platforms underrepresent $s_4$ ($2$--$7\%$), consistent with selection toward share-worthy, more articulate prompts.

\subsection{Topic separability}
\label{sec:r3}

If articulation segments were merely a byproduct of topic, we would expect strong segment\,$\leftrightarrow$\,topic dependence. We test this by re-tagging every prompt with our unified $7$-class taxonomy and computing Cramér's $V$ and mutual information between $s(p)$ and the topic label.

Pooled across all corpora ($n{=}16{,}447$), Cramér's $V{=}0.243$ (mutual information $0.130$), a weak-to-moderate association \citep{cohen1988statistical}. On WildChat alone, $V{=}0.255$; on LMSYS, $V{=}0.247$. Under the stricter re-extraction of Appendix~\ref{sec:appendix-rerun} the dependence falls further, to $V{=}0.150$. The aggregate number, however, hides real heterogeneity across topics.

Within-topic re-clustering (Table~\ref{tab:within-topic}, Appendix~\ref{sec:appendix-within-topic}) refits the GMM inside each topic and asks whether it recovers the pooled partition; since a single restart is not stable at this sample size, we report the mean over $10$ restarts. Six of seven topics land at $\overline{\mathrm{ARI}}{\geq}0.49$, so inside those the articulation types are essentially the pooled ones. The exception is H2 (Personal Decisions) at $0.31$ --- and H2 is $53\%$ of the sample.

The reading is \emph{conditional separability}. Articulation is not a relabelling of subject matter: the association with topic is weak and in six of seven topics the same types reappear inside the topic. But in the largest, bulk-advice topic the prompts spread fairly evenly across five segments (Table~\ref{tab:within-topic-distribution}) and admit several near-equivalent partitions. Concentration drives recoverability: H7 is $40\%$ $s_4$ and recovers at $0.68$, H6 is $46\%$ $s_5$ at $0.55$, H2 has no dominant mode and sits at $0.31$ --- though under the stricter re-extraction H2 rises to $0.710$, so this instability too is partly a measurement property. This bears on the \emph{typological} reading of the segments, which we do not lean on, and not on Section~\ref{sec:impl}, which enters topic as a control covariate and never conditions on a within-topic refit. The low-ARI topic is in fact where $s_4$ is \emph{rarest} ($3\%$ of H2), so H2's instability cannot produce the $s_4$ effect; L.7 shows that deficit reproducing inside all seven topics.

\subsection{Association with LLM response characteristics}
\label{sec:impl}

The descriptive structure we have established (factors, segments) is only worth the trouble if articulation segments actually predict something about LLM behavior. Table~\ref{tab:segment-outcome-means} gives the population-level answer: the raw per-segment mean for six response outcomes, before any controls.

\textbf{These are response characteristics, not a quality score.} We do not claim a longer reply, or one with more options, is better: for a vague prompt a short reply may be right. The vector describes \emph{what the model did}, most of which is a defensible reaction to its input. Exactly one outcome is normative --- when a request is under-specified, asking is the appropriate move \citep{clark1989contributing, shaikh2024grounding} --- so our finding is not ``$s_4$ gets shorter answers'' but that $s_4$ gets neither an answer nor a question.

\begin{table*}[t]
\centering\small
\begin{tabular}{lrrrrrrr}
\toprule
& $s_0$ & $s_1$ & $s_2$ & $s_3$ & $s_4$ & $s_5$ \\
& baseline & open-asker & specifier & named-comp.\ & long-form/low-d.\ & vague/situated \\
$n$ & $2{,}418$ & $2{,}408$ & $4{,}016$ & $2{,}975$ & $1{,}852$ & $2{,}778$ \\
\midrule
Response length (tokens)         & 309 & 299 & \textbf{369} & 246 & 260 & \textbf{240} \\
Recommendation specificity ($1$--$5$) & 3.47 & 3.18 & 3.55 & 3.33 & 3.33 & \textbf{2.76} \\
$n$ options recommended          & 3.98 & 3.66 & 3.65 & 3.58 & \textbf{0.82} & 2.67 \\
Asked for clarification          & $6.7\%$ & $7.4\%$ & $6.8\%$ & $7.4\%$ & $8.1\%$ & $\textbf{11.9\%}$ \\
Gave a direct answer             & $67\%$ & $55\%$ & $59\%$ & $58\%$ & $\textbf{19\%}$ & $44\%$ \\
Search-attribute coverage        & $57\%$ & $34\%$ & $54\%$ & $54\%$ & $\textbf{14\%}$ & $21\%$ \\
\bottomrule
\end{tabular}
\caption{Per-segment means of six response outcomes, pooled across corpora. Raw means without controls; the regression-controlled analogue is in Table~\ref{tab:covariation} and Appendix~\ref{sec:appendix-regression}. Bold marks the most extreme cell per row.}
\label{tab:segment-outcome-means}
\end{table*}

Three contrasts stand out.

\paragraph{The articulation-poverty failure mode ($s_4$).}
$s_4$ ($11\%$ pooled, $16\%$ of WildChat) is where users write a lot but say little of substance (low F2 density, high F6 monologue). The thin output is partly mechanical: with few constraints to enumerate, the low option count ($\sim$$0.8$ vs.\ $4.0$) and search coverage ($14\%$ vs.\ $57\%$) are what sparse input should produce, and we do not read them as a failure. The diagnostic finding is what the model does \emph{instead}: faced with under-specification it neither answers ($19\%$ direct-answer vs.\ $67\%$) nor asks back (clarification $+1.4$pp, logit $+0.23$, $p{=}0.11$, indistinguishable from baseline). It produces a reply that neither resolves the request nor moves toward resolving it.

\paragraph{$s_5$ as the control condition: under-specification alone does not explain $s_4$.}
The deflationary reading of $s_4$ is that it is simply \emph{ambiguity}: too little information was given, so of course the model cannot answer, and articulation does no work. $s_5$ rules this out. $s_5$ is also under-specified, and by the natural measure \emph{more} so --- the lowest F1 centroid of all six segments ($-1.05$ against $-0.95$; Table~\ref{tab:centroids}), the lowest recommendation specificity ($2.76$), the second-lowest search coverage ($21\%$). On the ambiguity account the two should behave alike; instead $s_5$ is the \emph{only} segment drawing substantially more clarification ($11.9\%$ vs.\ $6.7\%$; logit $+0.77$, $p{<}10^{-11}$) while $s_4$ draws none.

What separates them is not how much information is missing but how it is packaged: $s_5$ states its vagueness in a short, situated, first-person ask, while $s_4$ buries a diffuse ask in a long declarative monologue. Clarification tracks the packaging, not the information content --- which is why an input-side \emph{articulation} measure rather than an ambiguity score predicts it, and the response is withheld exactly where under-specification is hardest to see. $s_5$ pays its own price: the model asks back yet its answers stay vague, consistent with sycophancy-style adjustment to disclosed context \citep{sharma2024sycophancy}.

A third contrast is milder: $s_1$ ($15\%$, low search-attribute language) gets mid-length replies but a sharp drop in search coverage ($34\%$ vs.\ $57\%$), shifting toward experience attributes instead.

\paragraph{Significance under controls.}
Regressing each outcome on segment dummies with topic (H1--H7), assistant model, $\log(\text{prompt length})$, and source corpus as controls (OLS for continuous, logistic for binary), the segment dummies are jointly significant (likelihood-ratio) on all $12$ outcomes; the largest effects are on response length ($R^2{=}0.257$), search coverage ($0.201$), and recommendation specificity ($0.186$) (Table~\ref{tab:covariation}, Appendix~\ref{sec:appendix-regression}). Joint significance of the segment \emph{block} is not the claim of interest --- at this $n$ it would be surprising if a six-way partition explained nothing --- so we read the result through the per-segment effect sizes in Table~\ref{tab:regression}. The joint test on clarification is carried by $s_5$ ($+0.77$), not by $s_4$, whose coefficient stays indistinguishable from baseline ($+0.23$, $p{=}0.11$); the headline finding is that null, not the block test. Appendix~\ref{sec:appendix-nested} decomposes the block, showing segment membership adds significant power over topic, model, corpus, and length on every outcome.

Three confounds are ruled out separately. \emph{Length}: within every prompt-length quintile, including the shortest ($3$--$14$ tokens) where a long monologue is impossible, the direct-answer deficit holds at $29$--$48$pp (L.5). \emph{Topic}: $s_4$ concentrates nowhere (Table~\ref{tab:within-topic-distribution}), and refitting inside each topic group separately reproduces the deficit in all seven ($-19$ to $-54$pp, every $p{<}0.002$; L.7). \emph{Task contamination}: restricting $s_4$ to the $59\%$ of prompts containing a question mark preserves or intensifies every effect (L.3), and per-corpus refits reproduce the pattern in WildChat and LMSYS independently (L.1).

These remain \emph{associations}: with topic, model, length, and corpus controlled, we do not claim that, holding the user's latent goal fixed, articulating differently \emph{causes} a different response.

\subsection{Validation}
\label{sec:validation}

We audit the response-side measures with an inter-annotator study ($N{=}100$, $30$ from $s_4$): two annotators (one author, one paid external; see Ethical Considerations) re-coded six features in isolation, blind to the Stage~2 labels. The test is whether the extractor's disagreement with each annotator exceeds the disagreement between the two annotators themselves; if not, the extractor is within the natural inter-annotator band.

For five of six features (Table~\ref{tab:threeway-agreement}) the extractor's agreement with at least one annotator matches or exceeds the inter-annotator (A--B) baseline. The one exception (\texttt{gave\_direct\_answer}) reflects a rater's loose interpretation of the rubric: under the spec's strict rule, A--LLM $\kappa$ rises from $-0.02$ to $0.33$ overall and $0.58$ within $s_4$. The headline $s_4$ contrast on direct-answer rate and search coverage reproduces under both human annotators (Table~\ref{tab:threeway-contrast}).

Additional validation is in Appendix~\ref{sec:appendix-knowngroups}: known-groups directional checks ($4/4$ pass on $10$ synthetic prompts), self-consistency ($r{\geq}0.86$ for continuous features, $\kappa{=}0.88$ for the brand-explicitness label), and cross-corpus factor congruence (Table~\ref{tab:factors}, second column).

\paragraph{Reliability is not validity.} Those checks measure \emph{reliability}: the extractor is consistent with itself and moves as expected on constructed inputs. None of them, and no congruence coefficient, detects an extractor that is stably wrong. Two further checks address the input side: Appendix~\ref{sec:appendix-construct} establishes construct-level validity for five of six labels against a quantified noise floor, and Appendix~\ref{sec:appendix-rerun} re-runs Stage~1 with a stricter extractor, after which every stability statistic improves --- which a structure manufactured by extraction noise should not do. Neither substitutes for human validation of the Stage~1 features, which we lack (Limitations).

\paragraph{Topic-tag validation.}
We separately validated the Stage~5 topic taxonomy on the same sample: the two annotators relabeled each prompt's high-level topic (H1--H7) blind to the LLM tag. Human--human agreement is substantial ($\kappa{=}0.72$, raw $76\%$), and the tagger agrees with each human at $\kappa{=}0.61$ (raw $67\%$), the same inside-the-envelope pattern as the response-side audit, $\approx 0.1\,\kappa$ below the human baseline. Agreement concentrates on confident prompts (raw $89\%$ at self-reported confidence $\geq 4/5$); residual disagreement falls where the two humans also disagree. Per-class and confusion-matrix detail in Appendix~\ref{sec:appendix-robustness}, L.6.

\section{Discussion}
\label{sec:discussion}

\paragraph{Articulation as a measurable, structured axis.}
Where the prompt-sensitivity literature \citep{sclar2024sensitivity, mizrahi2024state} treats articulation variation as a nuisance to average over, our results show it has a few identifiable dimensions, unusually stable for a data-driven instrument ($\phi \geq 0.97$ cross-corpus on the four core factors), and is therefore \emph{addressable}: robustness benchmarks would benefit from coverage across articulation segments rather than across stochastic paraphrases.

\paragraph{Detection and benchmark coverage.}
Articulation-poor ($s_4$-equivalent) prompts are detectable without the full Stage~1 pipeline: raw word count gives $\text{AUC}{=}0.87$ and \texttt{specificity\_mean} alone gives $0.91$ (Appendix~\ref{sec:appendix-robustness}, L.4), so a deployed system can flag the at-risk population with a single threshold and route it to clarification handling. That is a feature of the result and our answer to whether a simpler method would have sufficed: for \emph{detection}, yes --- one threshold on one feature, which we recommend over our pipeline. What the factor analysis and clustering buy is what a threshold cannot: evidence that the flagged prompts are a reproducible population rather than a threshold artefact, an estimate of their share of real traffic, and the $s_4$/$s_5$ contrast showing the deficit is not plain ambiguity. Word count works as a \emph{cue} because length and density are correlated in chat, not as the cause (L.5, L.7). This matters for evaluation, which varies prompts almost only along topic and task \citep{liang2023helm, zheng2023mtbench, dubois2024alpacafarm}: hand-curated sets such as MT-Bench, AlpacaEval, Arena-Hard \citep{li2024alpacaeval}, and HELM sit in the clear, specific $s_0$/$s_1$ region, so the $s_4$ population is exactly what they miss.

\paragraph{From phenomenon to input-side structure.}
\citet{shaikh2024grounding} identify the clarification-deficit \emph{phenomenon} on hand-curated under-specified prompts but cannot say which naturalistic prompts trigger it, because theirs are constructed rather than sampled. We supply that: the at-risk population is defined not by topic, task, or surface length but by an articulation profile (very low F2 density plus long F6 monologue), and we give the first prevalence estimate ($16\%$ of WildChat) in real traffic. Nor is it defined by under-specification alone --- $s_5$ is at least as under-specified and draws clarification at nearly twice the base rate --- so the deficit responds to how the missing information is packaged, not how much is missing. Naming the structure turns the failure mode into a benchmark-stratification target and a prerequisite for remedies such as active preference elicitation \citep{li2023eliciting}. It also makes one intervention definable, which we flag as future work rather than a claim: an \emph{articulation-normalizing rewrite}, restating a prompt at a fixed point on the articulation axes before the model answers. Without an instrument for the axis there is no fixed point to rewrite to.

\paragraph{Equity implications.}
If articulation correlates with sociolinguistic factors (education, first-language status, expertise), the segment-level gaps would mean better-articulating users receive more thorough answers. Prior work documents lower-quality responses to non-native English speakers \citep{reusens2024native} and less-elaborated responses to African American English than to equivalent Standard American English \citep{zhou2025dialect}. Our segments are defined on articulation form, not speaker identity, and our corpora lack demographic labels \citep{bender2021stochastic}, so we cannot attribute the gap to any protected attribute; but the disparity is consistent with that evidence and is a natural target for follow-up.

\section{Conclusion}
\label{sec:conclusion}
How people phrase an advice-seeking prompt is not noise but a stable, low-dimensional structure that replicates across corpora and is largely separable from what they ask about. Treating articulation as a measurable property of the input, rather than a robustness nuisance, makes a previously invisible population legible: the long-form but information-poor prompts that draw shorter, vaguer answers and, tellingly, no clarifying question. We release the extracted structure as a measurement instrument and argue that LLM evaluation should stratify on articulation, not topic alone.

\section{Limitations}
\label{sec:limitations}

\textbf{Descriptive, not causal.} Chat logs identify the encoded prompt and the reply, never the user's latent intent $\theta$, so we make no causal claim of the form ``same intent, different articulation, different response''; that would require controlled multi-prompt elicitation. For the same reason, and because the corpora do not track users across conversations, our stability claims are at the level of the partition (bootstrap ARI) and factor structure (Tucker congruence), not the individual.

\textbf{Selection bias toward higher articulation.} The Stage~0 cascade admits $\sim$$0.93\%$ of prompts as advice-seeking and likely favors requests that articulate themselves \emph{explicitly}, so the analyzed corpus skews articulate and the segments are conditional on surviving this filter. The bias cuts \emph{against} our headline finding: the under-articulated $s_4$ population is the kind most likely to be dropped, so the $16\%$ $s_4$ prevalence in WildChat is best read as a conservative lower bound.

\textbf{Measurement validity.} A single LLM family produces both the articulation features (Stage~1) and the response outcomes (Stage~2), raising a shared-method-variance concern. We mitigate it four ways: the Stage~2 extractor sees only the reply, never the prompt; the inter-annotator audit (Section~\ref{sec:validation}) places it inside the human envelope on five of six features; the pre-registered known-groups experiment (Appendix~\ref{sec:appendix-construct}) shows five of six factors respond to targeted manipulations of their named constructs at $9\times$ to $49\times$ the extractor's test--retest noise; and all regressions control for length, topic, model, and corpus.

One inference we previously drew from the congruence figures needs correcting. Cross-corpus replication ($\phi \geq 0.97$) rules out a \emph{corpus-specific} artefact, not a \emph{model-specific} one, since the same extractor is applied to every corpus: an extractor that reliably but systematically under-counted intent spans in long narrative prompts would produce high congruence exactly as a valid one would. Appendix~\ref{sec:appendix-rerun} shows the structure sharpens rather than degrades under a stricter extractor, but that is not human ground truth on the input side, which stays the weakest link in the chain.

\textbf{Soft boundaries, anchored finding.} Silhouette is low ($0.19$) and bootstrap ARI moderate ($0.55$), so the segments are density modes in a continuous space, not discrete types; F5 and F6 also sit in the ``fair'' congruence band ($0.85$--$0.94$) rather than the ``equal'' band, and the within-topic refit in the largest topic is redrawable ($\overline{\mathrm{ARI}}{=}0.31$). We therefore do not defend the six segments as a typology. The $s_4$ finding survives that concession because it does not need one: it is anchored on F2 (density), one of the four core factors stable across corpora, its deficit is recoverable with no clustering at all from a single feature (\texttt{specificity\_mean}, $\text{AUC}{=}0.91$; L.4), and it reproduces inside every topic (L.7) and every length quintile (L.5).

\textbf{Scope.} The claim is bounded to English-language, first-turn, advice-seeking prompts across seven enumerated domains, drawn from three independently collected corpora and answered by six assistant platforms; healthcare is excluded because safety-rail behavior distorts responses there. Cross-corpus replication holds \emph{within} that subset. We claim nothing about non-English articulation, multi-turn conversations, or task-oriented prompts.

\section*{Ethical Considerations}

\textbf{Data sources and licensing.} All data are drawn from publicly released, de-identified chat corpora: WildChat \citep{zhao2024wildchat}, LMSYS-Chat-1M \citep{zheng2024lmsys}, and the ShareChat platforms \citep{yan2025sharechat}. WildChat is released under the ODC-BY License, LMSYS-Chat-1M under the LMSYS-Chat-1M Dataset License Agreement, and ShareChat under the CC BY-NC 4.0 License. Our use is consistent with the intended research purpose of each release. We do not collect any new user-generated data.

\textbf{Re-identification and privacy.} Example prompts shown in Appendix~\ref{sec:appendix-examples} are reproduced from the upstream corpora; where content could identify an individual we apply minimal additional redaction. Our analysis characterises articulation patterns at the population level; we do not attempt to identify, profile, or re-identify individual users.

\textbf{Annotator information.} The inter-annotator audit (Section~\ref{sec:validation}) was performed by one author and one paid external annotator. The external annotator was compensated at approximately \$10/hour for roughly $4$ hours of work, above the local minimum wage. The task involved reading publicly released chat conversations and applying a coding rubric. We determined that IRB review was not required because the annotation task processes publicly released, de-identified data rather than recruiting human subjects for original study; no protected attributes are elicited and no participant interaction is involved.

\section*{Acknowledgments}

This work was supported by the Institute of Information \& Communications Technology Planning \& Evaluation (IITP)---Global Data-X Leader HRD program grant funded by the Korea government (MSIT) (IITP-RS-2024-00440626), and by the KAIST C2 (Creative \& Challenging) Project (Project No.\ N11260031).

\section*{Use of AI Assistants}

We used AI assistants (Claude) for writing assistance (drafting and editing) and for code development. All final decisions on content, analysis, and claims were made by the authors. The commercial LLMs that form part of the empirical pipeline are not writing assistants and are documented as methodology in Section~\ref{sec:method}.

\bibliography{custom}

\appendix

\section{Candidate Feature List and Preprocessing}
\label{sec:appendix-features}

\subsection*{A.1 Raw feature set (Stage~1 output)}

The Stage~1 LLM extracts the following $30+$ raw features per prompt, before any preprocessing:

\begin{itemize}\setlength\itemsep{2pt}
\item \textbf{Span counts (12)}: \texttt{n\_spans}, \texttt{n\_search}, \texttt{n\_experience}, \texttt{n\_credence}, \texttt{n\_hard}, \texttt{n\_soft}, \texttt{n\_exemplar}, \texttt{n\_open}, \texttt{n\_brand}, \texttt{n\_quantitative}, \texttt{n\_self\_situation}, \texttt{n\_exemplar\_ref}.
\item \textbf{Compositional ratios (7)}: \texttt{r\_search}, \texttt{r\_experience}, \texttt{r\_credence} (Nelson attribute mix; sum to $1$); \texttt{r\_hard}, \texttt{r\_soft}, \texttt{r\_exemplar}, \texttt{r\_open} (Bettman hardness mix; sum to $1$).
\item \textbf{Categorical (2)}: \texttt{beta} (brand explicitness: absent/implied/named), \texttt{sentence\_mood} (declarative/interrogative/imperative/mixed).
\item \textbf{Continuous (9)}: \texttt{specificity\_mean}, \texttt{specificity\_var}, \texttt{crystallization}, \texttt{hedge\_count}, \texttt{hedge\_rate}, \texttt{imperative\_count}, \texttt{named\_alt\_count}, \texttt{token\_count}, \texttt{spans\_per\_token}.
\item \textbf{Person distribution (4)}: \texttt{first\_person\_count}, \texttt{second\_person\_count}, \texttt{third\_person\_count}, \texttt{first\_person\_rate}.
\end{itemize}

\subsection*{A.2 Preprocessing pipeline (raw $\to$ $19$ features)}

\begin{enumerate}\setlength\itemsep{1pt}
\item \textbf{ilr transform on compositional groups.} The Nelson 3-part attribute simplex maps to $\mathbb{R}^2$ (\texttt{ilr\_attr\_1}, \texttt{ilr\_attr\_2}); the Bettman 4-part hardness simplex maps to $\mathbb{R}^3$ (\texttt{ilr\_hard\_1}--\texttt{ilr\_hard\_3}). This breaks the sum-to-1 constraint and removes the spurious negative correlations that compositional data otherwise induce. After this step the seven raw ratios are replaced by five ilr coordinates.
\item \textbf{Length normalization.} All raw count features are length-normalized to rate variables: $\texttt{rate\_X} = \texttt{X}/\texttt{token\_count}$. This applies to \texttt{n\_spans}, \texttt{n\_brand}, \texttt{n\_quantitative}, \texttt{n\_self\_situation}, \texttt{n\_exemplar\_ref}, \texttt{imperative\_count}, \texttt{named\_alt\_count}, \texttt{hedge\_count}. The original counts are dropped.
\item \textbf{Drop near-zero-variance columns.} Features with variance below $10^{-4}$ after standardization are dropped. This removes redundant feature pairs (e.g., \texttt{hedge\_count} after \texttt{rate\_hedge\_count} is created), and several length-degenerate rates.
\item \textbf{Impute residual NaN.} Crystallization can be missing when the LLM output fails to populate the field; these cells ($<$$0.1\%$ of rows) are imputed by column mean.
\item \textbf{$z$-standardize.} Each remaining feature is centered and scaled to unit variance.
\end{enumerate}

The resulting matrix is $16{,}447 \times 19$. The 19 features that enter factor analysis are listed in Appendix~\ref{sec:appendix-loadings}. The categorical \texttt{beta} and \texttt{sentence\_mood} are not entered into the factor model; they are retained as descriptive variables and used in known-groups validation (Appendix~\ref{sec:appendix-knowngroups}).

\section{Topic Taxonomy and Annotation Guide}
\label{sec:appendix-topics}

The unified taxonomy adapted from \citet{yan2025sharechat}. Twenty-five fine-grained categories roll up to seven high-level groups:

\begin{itemize}\setlength\itemsep{2pt}
\item \textbf{H1 (Information \& News)}: factual\_question, knowledge\_explanation, news\_current\_events.
\item \textbf{H2 (Personal Decisions \& Recommendations)}: purchasable\_products, travel\_planning, food\_cooking, lifestyle\_health\_fitness, financial\_career\_advice.
\item \textbf{H3 (Education \& Tutoring)}: tutoring\_or\_teaching, homework\_help, skill\_learning.
\item \textbf{H4 (Technical / STEM)}: computer\_programming, mathematical\_calculation, scientific\_or\_engineering.
\item \textbf{H5 (Creative \& Media)}: creative\_writing, poetry\_lyrics, visual\_art\_design, music\_audio, roleplay\_fiction.
\item \textbf{H6 (Relational \& Emotional)}: relationships\_personal, mental\_emotional\_support, parenting\_family.
\item \textbf{H7 (Other \& General)}: argument\_or\_summary\_generation, legal\_civic, general\_other.
\end{itemize}

The original ShareChat taxonomy targets general chat; our analysis sample is the advice-seeking subset, so H2 is empirically the largest group ($n{=}8{,}730$, $53\%$ of pooled positives). The H1--H7 identifiers are retained verbatim from the original taxonomy for cross-reference.

\paragraph{Annotation procedure.} The full text of the guide given to the human annotators for the topic-tagging validation (Section~\ref{sec:validation}) is summarized here. Annotators labeled $100$ prompts, stratified to roughly $14$ per high-level class so each of the seven buckets has enough data for per-class agreement. Labeling was blind: the LLM tags were withheld until the annotator finished, which protects the agreement statistic from anchoring. For each prompt the annotator recorded (i) one high-level group ($\mathrm{H1}$--$\mathrm{H7}$), (ii) one of the $25$ fine-grained categories above, (iii) a confidence rating ($1$--$5$), and (iv) optional free-text notes for edge cases. The target pace was $20$--$30$ seconds per prompt.

\paragraph{Decision rules.} (1) Substance over phrasing: ``recommend a budget laptop'' and ``which laptop should I buy under \$800?'' are both H2 (\texttt{purchasable\_products}). (2) Pick one dominant category; record any alternative in notes. (3) Label the user's intent, not the response content: a prompt about a Python concept is H4 (\texttt{computer\_programming}), while H3 (\texttt{skill\_learning}) is reserved for prompts framed as learning paths (``I want to become a data scientist, what should I study?''). (4) Personal decision vs.\ abstract comparison distinguishes H2 (deciding for oneself) from H4 (an abstract technical comparison). (5) Exercise or diet prompts are H2 (\texttt{lifestyle\_health\_fitness}); medical or drug prompts were filtered at Stage~0 (healthcare exclusion) and, if one slips through, are marked H7 (\texttt{general\_other}) with a note. (6) ``Write a story about a dragon'' is \texttt{creative\_writing}, whereas ``you are a dragon, respond in character'' is \texttt{roleplay\_fiction}. (7) Summarization and pro/con or argument generation belong to H7 (\texttt{argument\_or\_summary\_generation}), because the user asks the system to \emph{produce} text rather than to be informed. (8) \texttt{general\_other} is the escape hatch, used only when no other category fits, always with a note.

\paragraph{Confidence scale.} $5$: certain, a single category unambiguously fits. $4$: confident, one category clearly dominant. $3$: lean, two categories plausible, one chosen. $2$: coin-flip between two fine categories within the same high-level group. $1$: genuinely ambiguous (prompt too short, garbled, or off-task). Ratings of $1$ or $2$ require a note.

\section{Segment Interpretations}
\label{sec:appendix-segments}

The $6$ articulation segments, with their dominant factors and training-set sizes ($n_\text{train}$ from WildChat $7{,}225$):

\begin{itemize}\setlength\itemsep{2pt}
\item \textbf{$s_0$ baseline} ($n_\text{train}{=}1{,}160$, $16\%$): mid-range on most factors; a generic articulation pattern.
\item \textbf{$s_1$ open-asker} ($n_\text{train}{=}1{,}048$, $15\%$): very low F4 (search-hard); open-ended advice prompts without specific spec requirements.
\item \textbf{$s_2$ specifier} ($n_\text{train}{=}1{,}881$, $26\%$): high F1 (specificity, ilr\_hard\_2/3); detailed constraint articulation. The largest segment.
\item \textbf{$s_3$ named-comparer} ($n_\text{train}{=}992$, $14\%$): high F2 (density), F3 (named-alt), F4 (search-hard) jointly; dense brand-aware spec-driven prompts.
\item \textbf{$s_4$ long-form / low-density} ($n_\text{train}{=}1{,}166$, $16\%$): very low F2 (sparse intent), very high F6 (long declarative monologue); verbose but information-poor.
\item \textbf{$s_5$ vague + situated} ($n_\text{train}{=}978$, $14\%$): very low F1 (vague), moderate F5 (some self-disclosure); the segment with the highest clarification rate.
\end{itemize}

Figure~\ref{fig:heatmap} (main text) shows the centroid loading profile per segment; numeric centroids are in Appendix~\ref{sec:appendix-centroids}.

\section{Full Factor Loadings}
\label{sec:appendix-loadings}

Table~\ref{tab:full-loadings} reports the full $19 \times 6$ Promax-rotated pattern matrix. Loadings $|\lambda|>0.30$ are bolded to highlight the dominant features per factor.

\begin{table*}[t]
\centering\small
\begin{tabular}{lrrrrrr}
\toprule
Feature & F1 & F2 & F3 & F4 & F5 & F6 \\
\midrule
ilr\_attr\_1            & $+0.05$ & $+0.01$ & $-0.03$ & \textbf{$+0.58$} & $-0.02$ & $+0.02$ \\
ilr\_attr\_2            & \textbf{$+0.54$} & $+0.14$ & $-0.05$ & $-0.10$ & $-0.10$ & $-0.09$ \\
ilr\_hard\_1            & $-0.11$ & $-0.09$ & $-0.04$ & \textbf{$+0.86$} & $+0.07$ & $+0.02$ \\
ilr\_hard\_2            & \textbf{$+0.72$} & $+0.18$ & \textbf{$-0.41$} & $-0.05$ & $-0.02$ & $+0.15$ \\
ilr\_hard\_3            & \textbf{$+0.62$} & $-0.16$ & $+0.15$ & $-0.05$ & $+0.10$ & $+0.01$ \\
rate\_n\_spans          & $+0.15$ & \textbf{$+0.90$} & $+0.18$ & $+0.00$ & $+0.09$ & $-0.08$ \\
rate\_n\_brand          & $+0.02$ & $+0.25$ & \textbf{$+0.62$} & $-0.02$ & $-0.06$ & $+0.13$ \\
rate\_n\_quantitative   & $+0.21$ & $+0.16$ & $-0.09$ & \textbf{$+0.38$} & $+0.10$ & $-0.01$ \\
rate\_n\_self\_situation & \textbf{$-0.42$} & \textbf{$+0.32$} & $-0.05$ & $+0.17$ & \textbf{$+1.01$} & $+0.22$ \\
rate\_n\_exemplar\_ref  & $+0.02$ & $-0.02$ & \textbf{$+0.68$} & $-0.07$ & $-0.01$ & $-0.11$ \\
rate\_imperative\_count & $-0.01$ & $+0.05$ & $-0.07$ & $+0.01$ & $-0.07$ & \textbf{$-0.55$} \\
rate\_named\_alt\_count & $-0.04$ & $+0.27$ & \textbf{$+0.73$} & $-0.03$ & $-0.07$ & $+0.18$ \\
rate\_hedge\_count      & $+0.05$ & $-0.01$ & $+0.00$ & $-0.10$ & $+0.09$ & $+0.04$ \\
specificity\_mean       & \textbf{$+0.65$} & $+0.07$ & $+0.27$ & $+0.23$ & $+0.02$ & $-0.09$ \\
specificity\_var        & \textbf{$+0.46$} & $+0.17$ & $-0.08$ & $-0.05$ & $-0.10$ & $+0.06$ \\
crystallization         & \textbf{$-0.50$} & $+0.22$ & $-0.11$ & $-0.28$ & $+0.03$ & $-0.07$ \\
first\_person\_rate     & $+0.10$ & $-0.04$ & $-0.06$ & $-0.02$ & \textbf{$+0.31$} & $-0.12$ \\
spans\_per\_token       & $+0.13$ & \textbf{$+0.89$} & $+0.20$ & $+0.00$ & $+0.07$ & $-0.05$ \\
token\_count            & $+0.11$ & \textbf{$-0.55$} & $-0.08$ & $+0.04$ & $-0.15$ & \textbf{$+0.56$} \\
\bottomrule
\end{tabular}
\caption{Full Promax-rotated factor loading matrix ($19$ features $\times$ $6$ factors). Bold: $|\lambda|>0.30$. Pattern matrix from \texttt{factor\_analyzer} with Promax rotation; sample size $n{=}16{,}447$. F1 = specificity/structure, F2 = density, F3 = named-alternative, F4 = search-attribute hardness, F5 = self-situation, F6 = length vs imperative. Note: \texttt{rate\_hedge\_count} has no factor loading $|\lambda|>0.3$ on its own: hedging covaries with self-situation (F5) and is captured at the feature level rather than as a standalone factor under the corrected preprocessing pipeline (see Section~\ref{sec:s3} on hedge feature deduplication).}
\label{tab:full-loadings}
\end{table*}

\begin{table}[t]
\centering\small
\begin{tabular}{lrr}
\toprule
Factor & Train/test $\phi$ & WC\,$\leftrightarrow$\,LM $\phi$ \\
\midrule
F1 (specificity)    & 0.986 & 0.973 \\
F2 (density)        & 0.995 & 0.982 \\
F3 (named-alt)      & 0.993 & 0.985 \\
F4 (search-hard)    & 0.990 & 0.968 \\
F5 (self-situation) & 0.904 & 0.819 \\
F6 (length-imp.)    & 0.890 & 0.811 \\
\bottomrule
\end{tabular}
\caption{Tucker's congruence $\phi$ per factor, summarised in Section~\ref{sec:r1}. \emph{Train/test}: single 50/50 split. \emph{WC\,$\leftrightarrow$\,LM}: WildChat vs.\ LMSYS. The four core factors clear the conventional $0.95$ ``equal'' threshold \citep{lorenzoseva2006tucker}; F5 and F6 sit in the ``fair'' band. Appendix~\ref{sec:appendix-rerun} calibrates these values against a permutation null and reports them under a stricter extractor.}
\label{tab:factors}
\end{table}

\paragraph{Loadings above $1.0$ under oblique rotation.} Two features (\texttt{rate\_n\_self\_situation} on F5, $+1.01$/$+1.02$) carry pattern-matrix loadings slightly above $1.0$. Under an oblique (Promax) rotation the pattern matrix holds partial regression weights, not correlations, so values above $1.0$ are admissible. Of the $19$ communalities, $18$ lie in $(0,1)$; \texttt{rate\_n\_self\_situation} alone yields a pattern-matrix communality of $1.39$. In oblique rotation, communalities can exceed $1$ because non-orthogonal factors contribute overlapping variance, and the quantity does not carry the same meaning as in orthogonal rotation. F5's $\lambda{=}1.02$ on this feature is a marker-variable signature of the underlying self-situation structure, not a degenerate solution.

\section{LLM Prompt Templates}
\label{sec:appendix-prompts}

We reproduce the operative parts of each LLM prompt used. The full templates and their version history are in the project repository.

\subsection*{E.1 Intent classifier (Stage~0 second pass)}
\begin{quote}\small\itshape
You read the user's FIRST message to an AI assistant and decide whether the user is asking for ADVICE, RECOMMENDATION, or DELEGATION about a real personal decision --- ideally one where the user has stated (or clearly implied) some preference, constraint, budget, situation, taste, or context.

Judge MEANING, not surface phrasing. The user does NOT have to use words like ``recommend'', ``suggest'', or ``advice''. Implicit and indirect expressions count when the intent to seek help on a decision is clear.

\textbf{YES} if the user is asking the assistant to help them: pick / choose / decide between options; recommend a product, brand, item, model, service, place, route, plan, action; suggest something to try / buy / read / watch / play / cook / visit / apply to; advise on a personal situation; plan / itinerate / strategize; verify a decision; compare named alternatives.

\textbf{NO} if the user is: issuing a task instruction (write code, summarize, translate, generate text, roleplay, continue a story); asking a pure factual / knowledge / how-to question with no personal stake; defining a persona / system prompt; providing synthetic-data templates; asking about the assistant itself; building a recommendation system or related engineering work.

Output EXACTLY one token: \textbf{YES} or \textbf{NO}.
\end{quote}

\subsection*{E.2 Span tagger (Stage~1)}
\begin{quote}\small\itshape
A ``span'' is a contiguous chunk of text expressing one preference, constraint, limitation, requirement, context fact, or self-situation. Per-span fields:
\textbf{attr\_type} (search/experience/credence/null) per Nelson;
\textbf{hardness} (hard/soft/exemplar/open/null) per Bettman;
\textbf{is\_brand} (bool);
\textbf{specificity} ($1$--$5$, $1$=vague, $5$=precise quantitative/named);
\textbf{quantitative} (bool);
\textbf{exemplar\_ref} (bool);
\textbf{self\_situation} (bool).
\\
Top-level fields:
\textbf{crystallization} ($1$--$5$, $1$=settled, $5$=exploratory);
\textbf{hedge\_markers} (list of verbatim hedge tokens);
\textbf{sentence\_mood} (interrogative/declarative/imperative/mixed);
\textbf{person\_distribution} (object with first/second/third counts);
\textbf{imperative\_count}; \textbf{named\_alternatives\_count}.
\\
Output a single JSON object with field \texttt{spans} (list) and the top-level fields. No prose, no markdown fences.
\end{quote}

\subsection*{E.3 Response profile extractor (Stage~2)}
\begin{quote}\small\itshape
For the assistant's response, extract:
\textbf{n\_recommended\_items} (int, distinct options recommended);
\textbf{recommended\_items} (list of short labels);
\textbf{clarifying\_questions} (list of questions asked back);
\textbf{attribute\_coverage} (object with three booleans: did the response discuss search / experience / credence attributes?);
\textbf{recommendation\_specificity} ($1$--$5$, null if no recommendation);
\textbf{hedging\_phrases} (verbatim list);
\textbf{gave\_direct\_answer} (bool);
\textbf{recommendation\_diversity\_kind} (narrow / moderate / diverse / n/a).
\\
Output a single JSON object. No prose, no markdown fences.
\end{quote}

\subsection*{E.4 Topic classifier (Stage~5)}
\begin{quote}\small\itshape
Classify the user's first message into one of $25$ fine-grained topic categories rolled up to seven high-level buckets (H1--H7; see Appendix~\ref{sec:appendix-topics}). Use the category that best describes the SUBSTANTIVE topic of the user's request, regardless of how the user phrased it. Output JSON with \texttt{fine}, \texttt{high\_level}, \texttt{confidence} ($1$--$5$), \texttt{rationale} ($<$$15$ words).
\end{quote}

\section{Segment Centroids in Factor Space}
\label{sec:appendix-centroids}

Numeric centroid coordinates of the $6$ articulation segments on WildChat training data ($n{=}7{,}225$, factor scores standardized to mean $0$, variance $1$ on the full sample):

\begin{table}[t]
\centering\footnotesize
\setlength{\tabcolsep}{3.5pt}
\begin{tabular}{lrrrrrr}
\toprule
seg & F1 & F2 & F3 & F4 & F5 & F6 \\
\midrule
$s_0$ baseline      & $+0.34$ & $+0.47$ & $+0.29$ & $+0.11$ & $+0.13$ & $+0.10$ \\
$s_1$ open-asker    & $+0.21$ & $-0.04$ & $-0.26$ & \textbf{$-1.20$} & $+0.42$ & $+0.11$ \\
$s_2$ specifier     & \textbf{$+1.00$} & $-0.32$ & $-0.29$ & $+0.40$ & $+0.23$ & $+0.35$ \\
$s_3$ named-comp.\  & $-0.12$ & \textbf{$+0.76$} & \textbf{$+0.87$} & \textbf{$+0.85$} & \textbf{$-0.89$} & \textbf{$-0.58$} \\
$s_4$ long-form     & \textbf{$-0.95$} & \textbf{$-1.19$} & $-0.17$ & $-0.28$ & \textbf{$-0.93$} & \textbf{$+1.24$} \\
$s_5$ vague/sit.\   & \textbf{$-1.05$} & $-0.27$ & $-0.23$ & $-0.33$ & \textbf{$+0.51$} & $-0.02$ \\
\bottomrule
\end{tabular}
\caption{Segment centroids on the WildChat training set under the corrected $k{=}6$ factor model. Bold: $|\text{score}|>0.5$. Compare with Figure~\ref{fig:heatmap}.}
\label{tab:centroids}
\end{table}

\section{Stage~0 Cascade Detail}
\label{sec:appendix-stage0}

\subsection*{G.1 Seven domain matchers}
Each matcher fires when either a strong keyword OR a regex pattern matches the prompt head ($\leq$$400$ characters). Excerpts of the keyword lists (full lists in \texttt{multi\_domain\_analyzer.py} and \texttt{src/domains.py}):

\begin{itemize}\setlength\itemsep{2pt}
\item \textbf{finance}: \emph{loan, mortgage, credit score, investment, 401k, IRA, retirement, tax return, ETF, mutual fund, financial advisor, debt consolidation, cryptocurrency} (29 strong terms + 5 patterns + 2 exclusions).
\item \textbf{shopping}: \emph{purchase, coupon, e-commerce, Amazon, eBay, Walmart} plus pattern \texttt{(?:recommend|\allowbreak suggest|\allowbreak best)\textbackslash s+\allowbreak(?:\textbackslash w+\textbackslash s+)\{0,3\}\allowbreak(?:product|\allowbreak phone|\allowbreak laptop|\allowbreak camera|\dots)}.
\item \textbf{travel}: \emph{flight, hotel, hostel, Airbnb, itinerary, visa, passport} plus patterns for \texttt{plan(ning)?\textbackslash s+a\textbackslash s+trip} and similar.
\item \textbf{career}: \emph{resume, cover letter, CV, job interview, salary negotiation, career change, promotion}.
\item \textbf{content\_media}: \emph{TV show, video game, board game, audiobook, playlist, podcast} plus patterns requiring a recommendation verb anchored to media (e.g., \texttt{recommend\allowbreak\textbackslash s+\dots\allowbreak(?:movies?|\allowbreak films?|\allowbreak series|\allowbreak books?|\allowbreak games?)}).
\item \textbf{lifestyle\_food}: \emph{recipe, meal, diet, workout, yoga, headphones, mattress, supplement}.
\item \textbf{advice\_generic} (pseudo-domain, recall-focused): \emph{any recommendation(s), any advice, what should I do, give me advice, any tips} plus patterns matching \texttt{any\textbackslash s+\dots\allowbreak\textbackslash s+\allowbreak(?:recommendations|\allowbreak suggestions|\allowbreak advices|\allowbreak tips|\allowbreak ideas)}.
\end{itemize}

\subsection*{G.2 e5-large advice anchors (verbatim)}
The $23$ anchor prototypes embedded against every non-domain prompt with \texttt{intfloat/e5-large-v2}:

\begin{enumerate}\setlength\itemsep{0pt}\small
\item I'm trying to decide whether to switch to a Roth IRA.
\item Should I take the new job offer or stay where I am?
\item I've been thinking about getting a new laptop for video editing under \$1500.
\item Can you help me figure out a vacation in Southeast Asia for 10 days?
\item I want to start lifting weights but my back has been bad --- what should I try?
\item Looking for a good board game for a group of six casual players.
\item Trying to pick between two mortgage offers --- 30yr at 7.1\% vs 15yr at 6.4\%.
\item What's a good restaurant in midtown for an anniversary dinner under \$200?
\item Any pizza recommendations?
\item Do you have any indie games to recommend for me?
\item Tell me a thing I should add to my bucket list.
\item Hello. Can you give me some advice to make peace with my partner?
\item What's a good website for me?
\item List romance films with strong female leads.
\item Give me a daily plan to learn faster.
\item Help me pick a gift for my dad.
\item I'm bored, what hobby should I pick up?
\item What course should I take in 2025?
\item Best candle with masculine smell?
\item Should I back up my data on USB sticks?
\item How can I make peace with my coworker?
\item X vs Y for someone like me?
\item Should I buy this or that?
\end{enumerate}

The threshold $\tau{=}0.80$ was calibrated on a held-out sample of $1{,}000$ rejected prompts: at $\tau{=}0.80$ the embedding net admits the top $\approx$$20$--$25\%$ of non-domain prompts, which empirically captures most of the implicit / underspecified advice prompts that the lexical matcher misses while excluding the long tail of unrelated content (e5 has high baseline similarity, $p_{50}{\approx}0.77$, so a low threshold admits essentially everything). Recall on a $500$-prompt FN sample is $98.4$--$99.6\%$ across corpora.

\section{Within-Topic Segment Distribution}
\label{sec:appendix-within-topic}

Cross-tabulation of segment by high-level topic. Row percentages report the segment distribution \emph{within} each topic group:

\begin{table}[t]
\centering\small
\setlength{\tabcolsep}{4pt}
\begin{tabular}{lrrr}
\toprule
Topic group & $n$ & $\overline{\mathrm{ARI}}$ & sd \\
\midrule
H7 (Other \& General)       & 1{,}935 & 0.684 & 0.025 \\
H5 (Creative \& Media)      & 1{,}015 & 0.582 & 0.028 \\
H6 (Relational \& Emotional)& 2{,}029 & 0.553 & 0.096 \\
H4 (Technical / STEM)       & 622    & 0.531 & 0.041 \\
H3 (Education)              & 1{,}257 & 0.518 & 0.071 \\
H1 (Information \& News)    & 859    & 0.494 & 0.039 \\
\midrule
H2 (Personal Decisions)     & 8{,}730 & \textbf{0.312} & 0.038 \\
\bottomrule
\end{tabular}
\caption{Within-topic re-clustering. The GMM is re-fit on each topic subset and its labels compared to the pooled-fit labels by ARI; we report the mean and sd over $10$ random restarts, since a single restart is not stable (H3, for instance, ranges $0.40$--$0.66$ across seeds). Values near $1$ mean the pooled segment structure recovers inside the topic. Six topics recover at $\overline{\mathrm{ARI}}{\geq}0.49$; only H2, the bulk-advice topic and $53\%$ of the sample, is substantially redrawable, and Table~\ref{tab:within-topic-distribution} explains why: H2 is the one topic with no dominant articulation mode.}
\label{tab:within-topic}
\end{table}

\begin{table*}[t]
\centering\small
\begin{tabular}{lrrrrrrrr}
\toprule
Topic & $n$ & $s_0$ & $s_1$ & $s_2$ & $s_3$ & $s_4$ & $s_5$ & top seg \\
 &  & baseline & open-asker & specifier & named-comp.\ & long-form & vague/sit.\ & \\
\midrule
H1 (Information \& News)     &   859  & 15\% & 11\% & 18\% & 18\% & \textbf{29\%} & \phantom{0}9\% & $s_4$ \\
H2 (Personal Decisions)      & 8{,}730 & 18\% & 14\% & \textbf{28\%} & 24\% & \phantom{0}3\% & 14\% & $s_2$ \\
H3 (Education \& Tutoring)   & 1{,}257 & 13\% & 18\% & \textbf{32\%} & 17\% & \phantom{0}8\% & 11\% & $s_2$ \\
H4 (Technical / STEM)        &   622  & 13\% & 10\% & \textbf{37\%} & 10\% & 20\% & 10\% & $s_2$ \\
H5 (Creative \& Media)       & 1{,}015 & 15\% & 18\% & \textbf{26\%} & 11\% & 24\% & \phantom{0}6\% & $s_2$ \\
H6 (Relational \& Emotional) & 2{,}029 & \phantom{0}7\% & 22\% & 13\% & \phantom{0}5\% & \phantom{0}7\% & \textbf{46\%} & $s_5$ \\
H7 (Other \& General)        & 1{,}935 & \phantom{0}8\% & 10\% & 15\% & 10\% & \textbf{40\%} & 17\% & $s_4$ \\
\midrule
\emph{pooled} $n$ per segment & 16{,}447 & 2{,}418 & 2{,}408 & 4{,}016 & 2{,}975 & 1{,}852 & 2{,}778 & \\
\bottomrule
\end{tabular}
\caption{Topic $\times$ segment cross-tab (row percentages within each topic); the last row gives the segment marginals, which the row percentages reproduce exactly. The specifier $s_2$ is modal in four topics, including the bulk-advice topic H2. Two topics have a strongly dominant, topic-specific mode: H6 is $46\%$ $s_5$ (vague and self-disclosing, unsurprising for relational and emotional problems) and H7 is $40\%$ $s_4$ (long-form/low-density). That concentration is what makes the within-topic GMM in Table~\ref{tab:within-topic} easy to recover: H7 recovers at $\overline{\mathrm{ARI}}{=}0.68$ and H6 at $0.55$. H2, conversely, has no dominant mode --- its mass spreads over five segments, none above $28\%$ --- and it is exactly the topic whose within-topic partition is redrawable ($0.31$).}
\label{tab:within-topic-distribution}
\end{table*}

Two features of this table matter for the argument. First, the association is real but local: $s_5$ concentrates in H6 and $s_4$ in H7, which is why $V{=}0.24$ is not zero. Second, and more consequentially, \textbf{$s_4$ is not a topic in disguise}. Its largest share is in H7 (Other \& General, $40\%$), a residual bucket rather than a subject matter, and it is present in every one of the seven groups --- including H4 Technical/STEM ($20\%$) and H5 Creative \& Media ($24\%$). Where it is \emph{rarest} is H2 ($3\%$), the single biggest topic. If $s_4$ were merely a proxy for a topic the model happens to be weak on, it would concentrate in that topic; instead it is spread thin across all seven, and its response deficit shows up separately inside each of them (Appendix~\ref{sec:appendix-robustness}, L.7).

\textbf{Note on H3 (Education).} Within the advice-seeking subset, H3 prompts are dominated by skill-learning and course/training-recommendation requests (``what should I learn to become a data scientist?'', ``recommend a beginner-friendly textbook for organic chemistry''). Pure homework-help prompts (``solve this differential equation'') are filtered out at Stage~0 because they are task instructions, not advice requests. The H3 prompts that remain therefore look more like H2-style decision-recommendation prompts in their articulation, which is why $s_2$ (specifier) is modal in both.

\section{Segment Regression Coefficients}
\label{sec:appendix-regression}

Coefficient table for the segment dummies on six representative outcomes. Reference category is $s_0$ (the modal baseline segment). Continuous outcomes use OLS on $\log(\text{outcome}+1)$ where appropriate (response length); binary outcomes use logistic regression with logit coefficients. All models include controls for topic (H1--H7), assistant model, $\log(\text{prompt token count})$, and source corpus.

\begin{table}[t]
\centering\small
\begin{tabular}{lrr}
\toprule
Outcome & $R^2$ & seg.\ LR $p$ \\
\midrule
Response length            & \textbf{0.257} & {<}0.001 \\
Coverage (search)          & 0.201 & {<}0.001 \\
Recommendation specificity$^\dagger$ & 0.186 & {<}0.001 \\
Coverage (count)           & 0.159 & {<}0.001 \\
Response hedge count       & 0.129 & {<}0.001 \\
Asked for clarification    & 0.114 & {<}0.001 \\
Gave direct answer         & 0.084 & {<}0.001 \\
Coverage (experience)      & 0.074 & {<}0.001 \\
Number of options          & 0.071 & {<}0.001 \\
Coverage (credence)        & 0.067 & {<}0.001 \\
Response hedge rate        & 0.058 & {<}0.001 \\
Clarification count        & 0.037 & {<}0.001 \\
\bottomrule
\end{tabular}
\caption{Covariation results (overall). $R^2$ is McFadden's pseudo-$R^2$ for binary outcomes, OLS $R^2$ for continuous. $p$ from likelihood-ratio test of joint segment dummies in the full regression. $n{=}16{,}447$ except where noted. $^\dagger$ Recommendation specificity is restricted to responses with at least one recommended item ($n{=}8{,}673$), since the outcome is undefined for replies that recommend nothing. The segment block is jointly significant on all twelve outcomes, which at this $n$ is a weak statement; the informative quantities are the per-segment coefficients in Table~\ref{tab:regression}, and in particular the $s_4$ clarification coefficient, which is \emph{not} significant ($p{=}0.11$) even though the block containing it is.}
\label{tab:covariation}
\end{table}

\begin{table*}[t]
\centering\small
\begin{tabular}{lrrrrrr}
\toprule
& Resp.\ length & Recomm.\ specif. & Coverage(search) & Asked clarify & Gave direct ans. & $n_\text{options}$ \\
& OLS coef & OLS coef & logit coef & logit coef & logit coef & OLS coef \\
\midrule
$s_1$ open-asker     & $-0.03$ & $-0.19^{***}$ & $-0.70^{***}$ & $+0.16$ & $-0.37^{***}$ & $-0.06$ \\
$s_2$ specifier      & $+0.01$ & $+0.08^{**}$  & $-0.04$ & $-0.00$ & $-0.18^{**}$  & $+0.03$ \\
$s_3$ named-comparer & $-0.04$ & $-0.02$ & $-0.25^{***}$ & $+0.25^{*}$ & $-0.59^{***}$ & $-0.70^{***}$ \\
$s_4$ long-form      & $-0.69^{***}$ & $-0.16^{**}$ & $-1.23^{***}$ & $+0.23$ & $-1.48^{***}$ & $-1.86^{***}$ \\
$s_5$ vague/situated & $-0.13^{***}$ & $-0.44^{***}$ & $-1.26^{***}$ & $+0.77^{***}$ & $-0.84^{***}$ & $-0.96^{***}$ \\
\bottomrule
\end{tabular}
\caption{Segment dummy coefficients relative to $s_0$ baseline. Significance: $^{*}p{<}0.05$, $^{**}p{<}0.01$, $^{***}p{<}0.001$ (Wald). For logistic outcomes the coefficient is a log-odds difference; for $\log(\text{response length}+1)$ a coefficient of $-0.69$ implies $\approx$$50\%$ shorter responses than baseline. Source corpus, model, topic, and prompt-length controls are not shown.}
\label{tab:regression}
\end{table*}

\textbf{Direction of effects.} The clearest pattern is on $s_4$ (long-form / low-density): substantially shorter responses ($e^{-0.69}-1 \approx -50\%$ on raw response length), much worse search coverage ($e^{-1.23} \approx 0.29$ odds vs baseline), no significant increase in clarification ($p{=}0.11$), far less direct answer given (logit $-1.48$, odds $0.23$), and far fewer options recommended ($-1.86$ on OLS, i.e.\ $\approx 1.9$ fewer named items). This is the segment of prompts where the user wrote a lot but said little of actionable substance; while the thinner content is partly expected given sparse input, the absence of any compensating rise in clarification (the model neither answers nor asks back) is the part that marks it as a failure mode rather than appropriate restraint.

The $s_5$ vague/situated segment is the \emph{only} one that triggers significantly more clarification ($+0.77^{***}$ on the logit), yet receives the lowest recommendation specificity ($-0.44$) and the second-lowest search coverage. The LLM \emph{recognises} the vagueness and asks back, but its substantive responses are still less specific. The $s_3$ named-comparer segment produces fewer options ($-0.70^{***}$) and lower direct-answer rate, consistent with the LLM treating brand-listing prompts as comparison requests rather than recommendation requests.

\section{Known-Groups Synthetic Prompts}
\label{sec:appendix-knowngroups}

The $10$ synthetic prompts used for construct-validity checking. Each prompt was hand-authored before any extraction was run; the predicted features for each prompt were then compared against the a-priori expected direction.

\begin{table}[t]
\centering\small
\begin{tabular}{lp{4.5cm}}
\toprule
Class & Prompt (excerpt) \\
\midrule
spec\_heavy & ``Looking for a 15-inch laptop, must have USB-C, 32GB RAM, dedicated GPU with at least 8GB VRAM, under \$1800, weight under 2.2kg. Should support Linux out of the box.'' \\
spec\_heavy & ``Recommend an SUV with at least 7 seats, AWD, towing capacity 5000lb+, MSRP under \$50k.'' \\
spec\_heavy & ``Need a wireless mouse, 800 DPI minimum, USB-C charging, must work with macOS and Linux, under \$80.'' \\
\midrule
experience\_heavy & ``I want a cozy cafe in Brooklyn for journaling on a rainy Saturday --- something with a warm aesthetic, soft lighting, maybe vintage.'' \\
experience\_heavy & ``Looking for a movie that feels like late autumn, melancholic but hopeful, with a slow burn.'' \\
experience\_heavy & ``Recommend a perfume that smells warm and grounded, like an old library with leather chairs.'' \\
\midrule
open\_explorer & ``I'm thinking of doing something different for my 30th. No idea what --- open to suggestions.'' \\
open\_explorer & ``What's a hobby I could pick up? I have no specific interest, just want to try something new.'' \\
\midrule
self\_situated & ``I'm 32, married, two young kids, household income \$130k, \$400k mortgage at 6.5\%, \$80k in 401k. Thinking about a Roth conversion --- does it make sense for someone like me?'' \\
self\_situated & ``Mid-career engineer at a 200-person startup, 8 years in, no manager track. Wondering if I should switch to a bigger company or stay and push for tech lead.'' \\
\bottomrule
\end{tabular}
\caption{Hand-authored known-groups prompts.}
\label{tab:knowngroups}
\end{table}

\subsection*{Inter-annotator audit tables (response-side features)}

Detail for the $N{=}100$ response-side audit described in Section~\ref{sec:validation}. Table~\ref{tab:threeway-agreement} reports pairwise agreement between the two human annotators and the Stage~2 extractor; Table~\ref{tab:threeway-contrast} shows that the headline $s_4$ contrast reproduces under both human annotators.

\begin{table}[t]
\centering\footnotesize
\setlength{\tabcolsep}{3pt}
\begin{tabular}{lccc}
\toprule
Feature & A--B & A--LLM & B--LLM \\
\midrule
\texttt{asked\_clarify} ($\kappa$)        & $0.74$ & $0.65$ & $0.52$ \\
\texttt{coverage\_search} ($\kappa$)      & $0.35$ & $0.34$ & $\mathbf{0.45}$ \\
\texttt{gave\_direct\_answer} ($\kappa$)  & $0.34$ & $-0.02$ & $0.18$ \\
\quad strict-rubric$^\dagger$             & ---    & $\mathbf{0.33}$ & --- \\
\midrule
\texttt{rec.\ specificity} ($\rho$)       & $0.67$ & $\mathbf{0.85}$ & $0.61$ \\
\texttt{n\_options} ($\rho$)              & $0.50$ & $0.43$ & $\mathbf{0.60}$ \\
\texttt{n\_clarify} ($\rho$)              & $0.77$ & $0.68$ & $0.51$ \\
\bottomrule
\end{tabular}
\caption{Pairwise agreement on the $N{=}100$ audit. A, B = two human annotators (one author + one paid external); LLM = Stage~2 extractor. Cohen's $\kappa$ (binary) / Spearman's $\rho$ (ordinal). \textbf{Bold}: extractor agreement with at least one annotator $\geq$ A--B baseline. $^\dagger$ Annotator A re-coded under the spec's strict rule; A--LLM $\kappa$ rises to $0.58$ within $s_4$.}
\label{tab:threeway-agreement}
\end{table}

\begin{table}[t]
\centering\footnotesize
\setlength{\tabcolsep}{3pt}
\begin{tabular}{lccc}
\toprule
$s_0 \to s_4$ rate    & Annot.\ A & LLM & Annot.\ B \\
\midrule
\texttt{gave\_direct\_answer} & $92{\to}93$ & $40{\to}20$ & $\mathbf{100{\to}43}$ \\
$\quad$($\Delta$ pp)          & ($+1$) & ($-20$) & ($\mathbf{-57}$) \\
\texttt{coverage\_search}     & $24{\to}13$ & $48{\to}7$ & $60{\to}30$ \\
$\quad$($\Delta$ pp)          & ($-11$) & ($-41$) & ($-30$) \\
\texttt{asked\_clarify}       & $0{\to}3$ & $4{\to}3$ & $0{\to}3$ \\
\bottomrule
\end{tabular}
\caption{Per-segment binary rates (\%) on the audit ($n_{s_0}{=}25$, $n_{s_4}{=}30$). The $s_4$ deficit on direct-answer and search-coverage reproduces under both human annotators and the extractor.}
\label{tab:threeway-contrast}
\end{table}

\textbf{Predicted features (means within class):} spec\_heavy prompts had $\bar{r}_\text{search}{=}0.94$ (vs $0.18$ experience\_heavy), $\bar{\text{specificity}}{=}4.75$ (vs $2.78$), crystallization $\bar{\,}{=}1.7$ (settled); open\_explorer crystallization $\bar{\,}{=}5.0$ (fully exploratory); self\_situated $\bar{\text{n\_self\_situation}}{=}3.5$ spans/prompt. All four a-priori directional predictions ($r_\text{search}$ on spec, specificity on spec, crystallization on open, self\_situation on self) were confirmed at the class-mean level. These prompts are small synthetic examples; they validate construct direction, not absolute accuracy on naturalistic data.

\subsection*{J.2 Self-consistency reliability}

Stage~1 features were re-tagged on a held-out $n{=}80$ sample with two independent LLM runs at temperature $0$ (different random seeds in the prompt header). Pairwise Pearson $r$ across continuous features ranges from $0.864$ (\texttt{n\_credence}) to $0.993$ (\texttt{specificity\_mean}); Cohen's $\kappa$ for the categorical brand-explicitness label is $\kappa{=}0.883$ (Table~\ref{tab:reliability}). This is a \emph{reliability} measure (same model, repeat runs), not a validity measure against human gold; the validity argument runs through the inter-annotator audit in Section~\ref{sec:validation}.

\begin{table}[t]
\centering\small
\begin{tabular}{lr}
\toprule
Feature & Pairwise $r$ \\
\midrule
specificity\_mean   & 0.993 \\
n\_experience       & 0.990 \\
crystallization     & 0.981 \\
n\_brand            & 0.980 \\
imperative\_count   & 0.975 \\
n\_spans            & 0.948 \\
n\_search           & 0.926 \\
hedge\_count        & 0.913 \\
n\_credence         & 0.864 \\
\midrule
$\beta$ (brand cat.) & $\kappa=0.883$ \\
\bottomrule
\end{tabular}
\caption{Self-consistency reliability ($n{=}80$, two independent runs).}
\label{tab:reliability}
\end{table}

\section{Example Prompts per Segment}
\label{sec:appendix-examples}

Three example prompts from each segment, from the WildChat training set, lightly truncated to $\sim$$180$ characters:

\begin{itemize}\setlength\itemsep{4pt}
\item \textbf{$s_0$ baseline} (compact mid-range):
\begin{itemize}\setlength\itemsep{1pt}
\item ``any legendary animes with good animation''
\item ``what is the best business to do with 1000 canadian dollars''
\item ``how do I find my people where we match interest, beliefs, values and vibe?''
\end{itemize}
\item \textbf{$s_1$ open-asker} (open-ended, low search-hard constraint):
\begin{itemize}\setlength\itemsep{1pt}
\item ``is pacing around my room bad for my joints? i try to raise my daily steps number by pacing back and forth around my room but i wonder if it could cause some health problems.''
\item ``I really care about my boyfriend but I don't feel like our relationship is healthy. What are some signs he's not the one for me?''
\item ``I got into a psychology major this year, but, the early and irregular weekly schedule and asocial nature of the course \dots'' (continues with open-ended uncertainty)
\end{itemize}
\item \textbf{$s_2$ specifier} (specific + hard-constraint):
\begin{itemize}\setlength\itemsep{1pt}
\item ``What websites sell alternative clothes in the UK that would suit a 32 year old man who mostly wears checked shirts with jeans and wants to dress more adventurously?''
\item ``Suggest a portfolio of ETFs with a hold time of 6-18 months based on the following geopolitical situation \dots''
\item ``recommend me material with links to free courses on how to start reverse engineering as a begginer''
\end{itemize}
\item \textbf{$s_3$ named-comparer} (dense + brands/exemplars):
\begin{itemize}\setlength\itemsep{1pt}
\item ``I am gonna build a website. Here are 4 domain names please suggest which is the best: captionsduniya.com / captionsbeast.com / Captionsbyte.com / captionsfactory.com''
\item ``recommend some star wars like movies (intergalactic, aliens, wisdom, etc)''
\item ``would it be cost effective or practical to run blender software on amazon web services?''
\end{itemize}
\item \textbf{$s_4$ long-form / low-density} (verbose context with diffuse intent: the user provides extensive background but the actionable ask is sparse):
\begin{itemize}\setlength\itemsep{1pt}
\item ``hi this is my info and you have to make a roadmap for my jee exams so that i can crack IIT under $1000$ rank \dots Age: 16, Location: Shahdol \dots Current Study Mode: Dummy school student \dots'' (long personal-context preamble, single broad ask)
\item ``I am running a hospital named Javitri hospital in India which is specialized in IVF \dots The main Doctor is Dr.\ Rajul Tyagi with 30 years of experience and 90\% success rate \dots One patient living in Canada contacted me via WhatsApp \dots'' (extensive business background, indirect request for advice)
\end{itemize}
\item \textbf{$s_5$ vague + situated} (personal context disclosure):
\begin{itemize}\setlength\itemsep{1pt}
\item ``I have 10 years of experience in web software development field. What can I do to improve my skill?''
\item ``I'm 60 years old and haven't achieved anything what I've missed out on.''
\item ``I am a published SF novelist, am fluent in German, have an MA in philosophy, have been a carer for six years within my family and am a medical herbalist with 24 years experience \dots''
\end{itemize}
\end{itemize}

These examples are reproduced as illustrations of within-segment articulation style. Personal information was minimally redacted only where it would identify an individual; otherwise the prompts are as collected by the upstream corpus.

\section{Reviewer-Anticipated Robustness Checks}
\label{sec:appendix-robustness}

We anticipate five robustness concerns and report the corresponding checks here. L.1 (pooling), L.3 (task contamination), L.5 (prompt length) and L.7 (topic competence) each remove one alternative explanation of the $s_4$ deficit; L.4 asks whether a simpler detector would do.

\subsection*{L.1 Per-corpus regression (pooling artifact check)}

The Implication section pools WildChat and LMSYS in the regression. To check that the segment effects are not a pooling artifact, we re-fit the regressions separately on each major corpus. Reference category is $s_0$; controls are $\log(\text{prompt tokens})$ only (topic and source dropped since per-corpus runs eliminate those confounders).

\begin{table}[t]
\centering\small
\begin{tabular}{lrr}
\toprule
Outcome & WildChat coef & LMSYS coef \\
& ($n{=}7{,}225$) & ($n{=}7{,}720$) \\
\midrule
Response length (log)       & $-0.91^{***}$ & $-1.11^{***}$ \\
Recommendation specificity  & $-0.32^{***}$ & $-0.25^{**}$ \\
Gave direct answer (logit)  & $-1.59^{***}$ & $-1.91^{***}$ \\
Search coverage (logit)     & $-2.07^{***}$ & $-1.68^{***}$ \\
\midrule
Asked clarification (logit) & $+0.09$ & $+0.83^{**}$ \\
\bottomrule
\end{tabular}
\caption{$s_4$ coefficient on each outcome, fit separately within each corpus (reference $s_0$; $\log$ prompt tokens is the only control, since a per-corpus fit removes the corpus and model confounds by construction). Significance: $^{**}p{<}0.01$, $^{***}p{<}0.001$. The four substantive deficits replicate independently in both corpora with the same sign and comparable magnitude, so the finding is not a pooling artefact. Clarification is the honest caveat and we separate it out: it is flat in WildChat ($p{=}0.67$), matching the pooled null, but in LMSYS $s_4$ \emph{does} draw somewhat more clarification ($p{=}0.001$), and the pooled estimate ($+0.23$, $p{=}0.11$) sits between them. We therefore state the claim as: $s_4$ shows no \emph{reliable} increase in clarification, in sharp contrast to $s_5$, whose increase is large and replicates everywhere. We do not claim clarification never rises for $s_4$ in any corpus.}
\label{tab:per-corpus}
\end{table}

\subsection*{L.2 Preprocessing audit and rerun}

An earlier draft of this analysis used a preprocessing matrix that contained a near-duplicate feature pair: \texttt{hedge\_rate} (computed at Stage~1) and \texttt{rate\_hedge\_count} (auto-generated by Stage~3 length-normalisation) both equal \texttt{hedge\_count}/\texttt{token\_count} and entered the factor model as numerically identical columns. The duplicate inflated a hedging factor to $|\lambda|>1.0$ on both columns. We patched preprocessing to drop one of the duplicates (and added a generic $|r|>0.9999$ duplicate check) and re-ran the full pipeline. The results reported in the main paper are from the corrected pipeline. The principal substantive shift is that the factor count drops from a noisy $k{=}7$ to a clean $k{=}6$ (hedging no longer factors out as a stand-alone dimension; it covaries with self-situation at the feature level), bootstrap ARI on segments drops from $0.78$ to $0.55$ (the duplicate had artificially sharpened cluster boundaries), and the $s_4$ long-form/low-density segment is preserved with essentially the same response-side deficits. The $s_3$ hedge-heavy segment from the earlier draft dissolves under the fix and is replaced by an $s_5$ vague/situated segment with the highest clarification rate.

\subsection*{L.3 $s_4$ with interrogative-only filter (task-contamination check)}

A reader could reasonably ask whether $s_4$ is contaminated by transformation/task prompts that slipped through Stage~0 (e.g., ``Respond to the following discussion post''). We restrict $s_4$ to the $59\%$ of $s_4$ prompts that contain a question mark, then compare against the $s_0$ baseline.

\begin{table}[t]
\centering\footnotesize
\setlength{\tabcolsep}{4pt}
\begin{tabular}{lrrr}
\toprule
Outcome & $s_0$ & $s_4$ (all) & $s_4$ (interrog.) \\
& $n{=}2{,}418$ & $n{=}1{,}852$ & $n{=}1{,}097$ \\
\midrule
Response length (tokens) & $309$ & $260$ & $\mathbf{231}$ \\
$n$ options              & $3.98$ & $0.82$ & $\mathbf{0.80}$ \\
Asked clarify            & $6.7\%$ & $8.1\%$ & $\mathbf{9.3\%}$ \\
Direct answer            & $66.7\%$ & $19.0\%$ & $\mathbf{20.3\%}$ \\
Search coverage          & $57.2\%$ & $14.0\%$ & $\mathbf{14.2\%}$ \\
Recomm.\ specificity     & $3.47$ & $3.33$ & $\mathbf{3.31}$ \\
\bottomrule
\end{tabular}
\caption{Restricting $s_4$ to prompts containing a question mark (\emph{interrog.}) preserves or intensifies every response-side effect. The long-form/low-density failure mode is not driven by task-like contamination.}
\label{tab:s4-interrogative}
\end{table}

\subsection*{L.4 Deployment-friendly proxy classifier (Q7)}

A practitioner who wants to detect articulation-poor prompts in production may not want to run the full Stage~1 LLM extraction. We test single-feature classifiers for $s_4$ membership.

\begin{table}[t]
\centering\footnotesize
\setlength{\tabcolsep}{4pt}
\begin{tabular}{lcrr}
\toprule
Proxy & LLM? & AUC & FPR@TPR$=0.80$ \\
\midrule
\texttt{token\_count}            & no  & $0.874$ & $0.116$ \\
\texttt{avg\_sentence\_length}   & no  & $0.586$ & $0.712$ \\
\texttt{question\_density}       & no  & $0.549$ & $0.563$ \\
combined (logit)                 & no  & $0.837$ & $0.219$ \\
\midrule
\texttt{specificity\_mean}       & yes & $0.908$ & $0.184$ \\
\texttt{spans\_per\_token}       & yes & $0.986$ & $0.027$ \\
\bottomrule
\end{tabular}
\caption{Single-feature ROC AUC for $s_4$ detection. \texttt{spans\_per\_token} ($0.986$) is essentially definitional --- F2 density is dominated by this feature --- so its AUC reflects how the segment was built, not a discovery. The deployment-relevant rows are the others: an LLM-free proxy, raw word count, reaches $0.874$, and a single Stage-1 feature, \texttt{specificity\_mean}, reaches $0.908$ with no factor model and no clustering. \textbf{This is also our answer to ``could a simpler method do this?''} For the \emph{detection} task, yes --- one threshold on one feature, and we recommend it. What the factor analysis and clustering buy is what a threshold cannot: evidence that the flagged prompts form a reproducible population, an estimate of how much of real traffic they are, and the $s_4$/$s_5$ contrast showing the deficit is not plain under-specification (Section~\ref{sec:impl}). The word-count AUC is high enough ($0.874$) that it is worth restating why length is a cue and not the cause: L.5 shows the response deficit persists inside every length quintile, and L.7 inside every topic.}
\label{tab:q7-proxy}
\end{table}

\subsection*{L.5 Matched-length analysis: density is not a length proxy}

A single LLM-free feature (raw token count) reaches $\text{AUC}{=}0.87$ for $s_4$ detection (L.4), which invites the objection that the $s_4$ response deficit is merely a length effect: longer prompts get worse responses, and $s_4$ is just the long-prompt segment. The regressions in Section~\ref{sec:impl} already control for $\log(\text{prompt length})$, but a linear control may not fully absorb a non-linear length effect. We therefore stratify into token-count quintiles and compare $s_4$ against the $s_0$ baseline \emph{within each quintile}, where prompt length is approximately held fixed.

\begin{table}[t]
\centering\small
\setlength{\tabcolsep}{4pt}
\begin{tabular}{lccccc}
\toprule
Quintile & $n$ ($s_0$/$s_4$) & \multicolumn{2}{c}{Direct (\%)} & $\Delta$ & Search (\%) \\
(tokens) & & $s_0$ & $s_4$ & (pp) & $s_0$/$s_4$ \\
\midrule
Q1 (3--14)    & 807/143 & 72.6 & 25.2 & $-47.4$ & 61.7/25.2 \\
Q2 (14--26)   & 732/34  & 65.7 & 17.6 & $-48.1$ & 57.8/8.8 \\
Q3 (26--96)   & 661/189 & 62.9 & 16.9 & $-46.0$ & 53.0/7.4 \\
Q4 (96--309)  & 172/680 & 62.8 & 21.6 & $-41.2$ & 54.1/16.6 \\
Q5 (309+)     & 46/806  & 45.7 & 16.1 & $-29.6$ & 43.5/11.7 \\
\bottomrule
\end{tabular}
\caption{Direct-answer rate and search-attribute coverage for $s_0$ vs.\ $s_4$ within token-count quintiles. The $s_4$ direct-answer deficit persists at $29$--$48$ percentage points in \emph{every} length stratum, including the shortest quintile (Q1, $3$--$14$ tokens) where a long monologue is structurally impossible. Length therefore does not explain the deficit: the gap is a property of articulation density, not prompt length. The crossover in $s_0$/$s_4$ sample sizes across quintiles also confirms the two segments are not simply long vs.\ short: both span the full length range.}
\label{tab:matched-length}
\end{table}

The $s_4$ deficit survives within every quintile (Table~\ref{tab:matched-length}). The most decisive stratum is Q1 ($3$--$14$ tokens): even among the shortest prompts (where the ``long-form monologue'' description cannot apply on length grounds), $s_4$ prompts receive a direct answer $25.2\%$ of the time vs.\ $72.6\%$ for $s_0$, a $47.4$pp gap comparable to the pooled effect. The search-coverage and option-count gaps behave the same way. Because the contrast holds with prompt length effectively fixed, the response deficit is attributable to low articulation density rather than to verbosity, and the $\text{AUC}{=}0.87$ of raw token count reflects that length and density are correlated in the wild, not that length is the operative cause.

\subsection*{L.6 Topic-tag validation against two human annotators}

To validate the Stage~5 topic taxonomy (cf.\ Section~\ref{sec:s5}), the two human annotators from the response-side audit (one author and one paid external annotator) re-labeled each of the $N{=}100$ audit prompts with the $7$-way high-level topic (H1--H7), blind to the LLM tag. All three label sets (A1, A2, LLM) are then cross-tabulated.

\begin{table}[t]
\centering\small
\begin{tabular}{lrr}
\toprule
Comparison        & raw \%      & Cohen $\kappa$ \\
\midrule
A1 vs.\ A2 (human--human) & $76.0$ & $\mathbf{+0.715}$ \\
A1 vs.\ LLM       & $67.0$      & $+0.615$ \\
A2 vs.\ LLM       & $67.0$      & $+0.614$ \\
\midrule
\multicolumn{3}{l}{\emph{three-way:}} \\
all three unanimous       & \multicolumn{2}{r}{$57 / 100$} \\
two humans agree (A1{=}A2) & \multicolumn{2}{r}{$76 / 100$} \\
all three different       & \multicolumn{2}{r}{$~4 / 100$} \\
\bottomrule
\end{tabular}
\caption{Pairwise agreement on the $7$-way high-level topic decision. Human--human $\kappa{=}0.72$ is the inter-annotator ceiling; the LLM tagger sits $\approx 0.1\,\kappa$ below it on each pairwise comparison.}
\label{tab:topic-agreement}
\end{table}

\begin{table}[t]
\centering\footnotesize
\setlength{\tabcolsep}{4pt}
\begin{tabular}{lccc}
\toprule
H class & support & F1$_{\mathrm{A1}}$ & F1$_{\mathrm{A2}}$ \\
\midrule
H4 Tech/STEM   & $14$/$17$ & $0.86$ & $\mathbf{0.90}$ \\
H6 Relational  & $13$/$14$ & $\mathbf{0.96}$ & $\mathbf{1.00}$ \\
H3 Education   & $14$/$17$ & $0.64$ & $0.77$ \\
H5 Creative    & $27$/$19$ & $0.68$ & $0.73$ \\
H2 Personal    & $12$/$23$ & $0.64$ & $0.51$ \\
H7 Other       & $15$/$\phantom{0}9$ & $0.41$ & $0.35$ \\
H1 Info/News   & $\phantom{0}5$/$\phantom{0}1$ & $0.42$ & $0.13^{\dagger}$ \\
\bottomrule
\end{tabular}
\caption{Per-class F1 with the LLM tag treated as a predictor and each human annotator's label treated as gold. \emph{support} = number of prompts (A1/A2); F1$_{\mathrm{A1}}$/F1$_{\mathrm{A2}}$ use annotator A1/A2 as gold. H4 and H6 are nearly perfectly recovered; H7 (Other) is the weakest class, reflecting that ``other-than-everything-else'' is itself an underdefined target. $^{\dagger}$H1 support of $1$ on A2 makes the F1 unstable.}
\label{tab:topic-perclass}
\end{table}

\begin{table}[t]
\centering\small
\setlength{\tabcolsep}{4pt}
\begin{tabular}{lrr}
\toprule
A1 confidence band & $n$ & A1 $=$ LLM (raw \%) \\
\midrule
Low  ($1$--$2$) & $29$ & $41.4$ \\
Mid  ($3$)      & $35$ & $65.7$ \\
High ($4$--$5$) & $36$ & $\mathbf{88.9}$ \\
\bottomrule
\end{tabular}
\caption{Agreement between annotator A1 and the LLM tag stratified by A1's self-reported confidence. When A1 finds the prompt unambiguous, LLM agreement reaches $89\%$; disagreement is concentrated on prompts where the annotator's own confidence is low.}
\label{tab:topic-confidence}
\end{table}

\begin{table}[t]
\centering\footnotesize
\setlength{\tabcolsep}{2.5pt}
\begin{tabular}{l|rrrrrrr}
\toprule
A1 $\downarrow$ \ LLM $\to$ & H1 & H2 & H3 & H4 & H5 & H6 & H7 \\
\midrule
H1 &  4 &  0 &  0 &  0 &  0 &  0 &  1 \\
H2 &  2 &  9 &  0 &  1 &  0 &  0 &  0 \\
H3 &  3 &  0 &  9 &  1 &  0 &  0 &  1 \\
H4 &  0 &  1 &  1 & 12 &  0 &  0 &  0 \\
H5 &  2 &  4 &  1 &  0 & 14 &  0 &  6 \\
H6 &  0 &  0 &  0 &  0 &  0 & 13 &  0 \\
H7 &  3 &  2 &  3 &  0 &  0 &  1 &  6 \\
\bottomrule
\end{tabular}
\caption{Confusion matrix, annotator A1 (rows) vs.\ LLM tag (columns). The dominant off-diagonal pattern is the LLM moving prompts from A1's H5 (Creative \& Media) and H7 (Other) into adjacent classes: the LLM under-recalls H5 ($14/27$ retained) and over-conflates H7 with H1/H2/H3. Both error modes are at the boundaries humans also find difficult.}
\label{tab:topic-confusion}
\end{table}

\paragraph{Fine-grained ($25$-way).} On the $25$-way fine-grained label, agreement drops as expected: A1--A2 $\kappa{=}0.70$ (raw $73\%$), A1--LLM $\kappa{=}0.55$ ($58\%$), A2--LLM $\kappa{=}0.57$ ($60\%$). The LLM--human gap is the same $\sim 0.13\,\kappa$ as on the high-level decision, indicating that the fine-grained tagger does not introduce additional systematic error beyond what the harder label set already imposes.

\paragraph{Summary.} The topic tagger sits inside the inter-annotator envelope ($\kappa{=}0.61$ vs.\ a human ceiling of $\kappa{=}0.72$). Within-class performance is excellent for the well-defined classes (H4, H6) and weak on the deliberately-residual H7 ``Other'' bucket, where humans themselves disagree. Crucially, the regressions in Section~\ref{sec:impl} treat topic as a control covariate rather than as a research variable, so this level of taxonomy noise does not threaten the headline finding: a tag that disagrees with humans roughly as often as humans disagree with each other still adsorbs the same topic-level variance from the response-outcome regressions.

\subsection*{L.7 $s_4$ deficit within each topic group}
\label{sec:appendix-topic-deficit}

Section~\ref{sec:impl} controls for topic as a covariate, which constrains the topic effect to be additive on the link scale. That is not enough to answer the sharper objection: models give worse answers on topics they know less well, and if $s_4$ prompts happened to concentrate in those topics, an ``articulation'' effect would be a topic-competence effect wearing a disguise. Table~\ref{tab:within-topic-distribution} already shows that $s_4$ does not concentrate --- it is spread across all seven groups and is \emph{rarest} in the largest one --- but the direct test is to refit the contrast inside each topic separately, with no pooling and no topic term at all.

\begin{table}[t]
\centering\footnotesize
\setlength{\tabcolsep}{3pt}
\begin{tabular}{lrrrrr}
\toprule
Topic & $n$ ($s_0$/$s_4$) & \multicolumn{2}{c}{Direct (\%)} & $\Delta$ & Search \\
 & & $s_0$ & $s_4$ & (pp) & $\Delta$ (pp) \\
\midrule
H1 Info/News   & 125/250  & 61.6 & \phantom{0}9.2 & $-52.4$ & $-41.6$ \\
H2 Personal    & 1600/225 & 70.8 & 44.9 & $-25.9$ & $-19.6$ \\
H3 Education   & 164/\phantom{0}99  & 64.0 & 19.2 & $-44.8$ & $-41.7$ \\
H4 Tech/STEM   & \phantom{0}82/127  & 56.1 & 33.9 & $-22.2$ & $-44.0$ \\
H5 Creative    & 150/244  & 64.7 & 10.2 & $-54.4$ & $-27.2$ \\
H6 Relational  & 146/133  & 44.5 & 25.6 & $-19.0$ & $+0.3$ \\
H7 Other       & 151/774  & 59.6 & 13.7 & $-45.9$ & $-19.8$ \\
\bottomrule
\end{tabular}
\caption{$s_4$ against the $s_0$ baseline, refit \emph{inside} each topic group: direct-answer rates and their gap, with the same gap on search-attribute coverage in the last column. Every direct-answer gap is negative and significant (Fisher exact, $p{<}0.002$ in all seven; the smallest effect is H6 at $-19.0$pp, $p{=}1.1\times10^{-3}$). The deficit is therefore not carried by any single topic and does not disappear when topic is held fixed by construction rather than by a regression term. The one exception anywhere in the table is search coverage in H6 (Relational \& Emotional), which is flat --- but H6 is the topic where search attributes barely apply at all ($s_0$ itself reaches only $10.3\%$), so there is no headroom for a gap.}
\label{tab:topic-deficit}
\end{table}

Table~\ref{tab:topic-deficit} reports the result. The direct-answer deficit reproduces in all seven topics, ranging from $-19$pp (H6) to $-54$pp (H5), each significant at $p{<}0.002$. Notably it is \emph{smallest} exactly where a topic-competence account predicts it should be largest --- H2, the bulk-advice topic the model sees most of --- and largest in H5 and H1. Together with the matched-length analysis (L.5), this closes the two obvious confounds: the $s_4$ response deficit survives holding topic fixed and holding prompt length fixed, separately.

\section{Model-Selection Curves}
\label{sec:appendix-modelsel}

\begin{table}[t]
\centering\small
\begin{tabular}{crr}
\toprule
$k$ & BIC & Silhouette \\
\midrule
2 & \phantom{0}54{,}592 & 0.197 \\
3 & \phantom{0}38{,}280 & 0.192 \\
4 & \phantom{0}35{,}313 & \textbf{0.215} \\
5 & \phantom{0}27{,}611 & 0.190 \\
\textbf{6} & \textbf{21{,}645} & 0.191 \\
\midrule
\multicolumn{3}{l}{\emph{beyond the factor-count cap:}} \\
7 & \phantom{00}{\phantom{-}}439 & 0.140 \\
8 & \phantom{00}{\phantom{-}}494 & 0.149 \\
9 & \phantom{0}$-3{,}589$ & 0.145 \\
10 & \phantom{0}$-4{,}865$ & 0.140 \\
\bottomrule
\end{tabular}
\caption{GMM $k$-selection curves on the WildChat training set ($n{=}7{,}225$, $6$-factor scores). Lower BIC is better; higher silhouette is better. Within the factor-count cap ($k \leq 6$) BIC decreases monotonically and selects $k{=}6$, while silhouette peaks at $k{=}4$. If the cap is lifted, BIC keeps decreasing through $k{=}10$ (no interior minimum), but silhouette drops sharply from $0.19$ at $k{=}6$ to $0.14$ at $k{=}7$ and stays flat thereafter. We read the post-$6$ BIC gains as statistical refinement of existing modes rather than the emergence of new interpretive types: the components added beyond $k{=}6$ do not separate cleanly (silhouette collapse), so they buy likelihood without distinctness. We therefore cap $k$ at the factor count ($6$), the largest value at which segments remain distinct enough for stable semantic labels. We follow BIC over silhouette within the capped range for the reason in Section~\ref{sec:s4}: silhouette assumes spherical equal-variance clusters, violated in the oblique-rotated factor space, while BIC is the likelihood-based criterion appropriate for Gaussian mixtures \citep{fraley2002mclust}. The bootstrap ARI for $k{=}6$ ($\overline{\,}{=}0.55$, Section~\ref{sec:r2}) confirms the partition is reproducible at the moderate level.}
\label{tab:k-selection-full}
\end{table}

\paragraph{Choosing the number of factors.} We use parallel analysis \citep{horn1965parallel}: we compute eigenvalues from $B{=}50$ random standard-normal draws of the same shape as the data matrix, take their mean as the random benchmark, and retain factors whose data eigenvalues exceed it. This accounts for the sampling noise that Kaiser's eigenvalue-$>$$1$ criterion ignores \citep{fabrigar1999evaluating}.

\paragraph{Factor-count robustness.} Parallel analysis retains $k{=}6$ under both random benchmarks. At $k{=}6$ the observed eigenvalue $1.042$ exceeds both the standard-normal (Horn) benchmark $1.026$ and the column-permutation benchmark $1.027$; at $k{=}7$ the observed $0.965$ falls below both. The factor count therefore does not depend on the parallel-analysis variant.

\begin{figure}[t]
\centering
\includegraphics[width=\columnwidth]{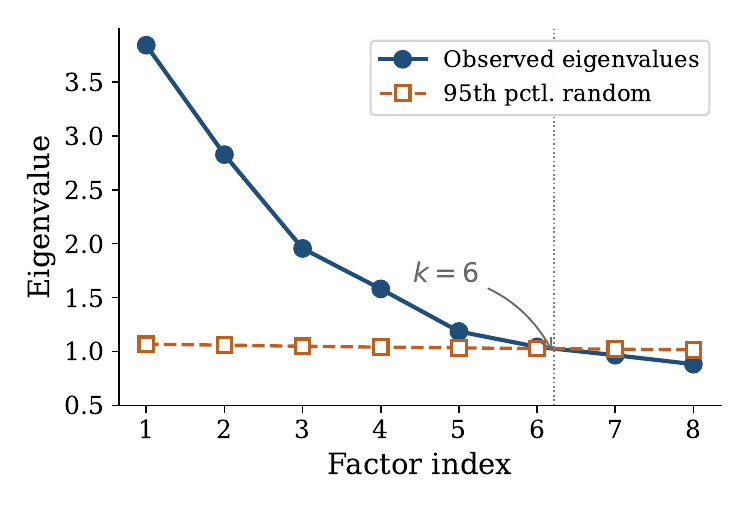}
\caption{Parallel-analysis scree plot. Solid: data eigenvalues. Dashed: mean random eigenvalues from standard-normal draws. The crossover at $k{=}6$ retains all factors with non-random variance.}
\label{fig:scree}
\end{figure}

\section{The Three Estimands, Stated Formally}
\label{sec:appendix-estimands}

Section~\ref{sec:claims} gives these in plain language; this is the formal statement. For a prompt $p$, segmentation assigns
\[
  s(p) \;=\; \operatorname*{arg\,max}_{k \in \{1,\dots,K\}} \Pr\!\big(k \mid a(p)\big),
\]
with $K{=}6$ fixed by the GMM (Section~\ref{sec:s4}).

\textbf{Structure.} Reproducibility of the partition is the bootstrap ARI between $s$ and a resampled partition $s^{(b)}$, reported as a mean and CI.

\textbf{Separability.} Is $s \perp z$? We report Cramér's $V$ and mutual information $I(s;z)$, and re-cluster within each topic: if the same articulation types reappear inside every topic, the structure is not a relabelling of subject matter.

\textbf{Covariation.} Does
\[
  \mathbb{E}\!\big[r \mid s, z, \mathrm{model}, \mathrm{len}\big] \;=\; \mathbb{E}\!\big[r \mid z, \mathrm{model}, \mathrm{len}\big]
\]
fail? We test it by regressing each outcome in $r$ on segment dummies together with topic, assistant model, source corpus, and $\log$ prompt length.

\section{Stricter Stage-1 Re-extraction and Calibration of the Congruence Statistic}
\label{sec:appendix-rerun}

The pipeline's inputs come from a single LLM extractor, so a natural worry is that its structure is an artefact of that extractor rather than of the data. The direct test is to swap the extractor for a stronger one applying a stricter standard and re-run everything. We re-ran Stage~1 with \texttt{claude-opus-4.8} in place of \texttt{claude-haiku-4-5}. The stricter extractor declines $1{,}241$ prompts ($7.5\%$) that the original extraction had admitted. We then re-executed Stages~3--6 end to end on the resulting matrix, refitting the factor model, the GMM, and every regression; nothing was carried over from the original run.

\begin{table}[t]
\centering\footnotesize
\setlength{\tabcolsep}{3pt}
\begin{tabular}{p{2.9cm}rr}
\toprule
 & main text & re-run \\
\midrule
Latent factors (parallel analysis) & 6 & 6 \\
Train/test Tucker $\phi$ & 0.890--0.995 & 0.993--0.9995 \\
WildChat\,$\leftrightarrow$\,LMSYS $\phi$ & 0.811--0.985 & 0.970--0.997 \\
Bootstrap ARI (segments) & 0.549 & 0.785 \\
Cramér's $V$ (segment $\times$ topic) & 0.243 & 0.150 \\
Within-topic ARI, H2 & 0.312 & 0.710 \\
\bottomrule
\end{tabular}
\caption{Every structural statistic improves under a stricter extraction standard, and the factor count is unchanged. In the main-text run, F5 and F6 sat at $\phi \approx 0.82$ cross-corpus and were hedged in Section~\ref{sec:r1} as partly dataset-specific; in the re-run all six factors clear $0.97$. A structure manufactured by extraction noise should degrade, not sharpen, when the extractor is made stricter.}
\label{tab:rerun}
\end{table}

\paragraph{Calibrating $\phi$ against a null.} Tucker's $\phi$ is a cosine and the conventional bands (``$>0.95$ = excellent'') are not calibrated against any distribution. We therefore built one. Permuting every feature column independently destroys all covariance while preserving every marginal; running the identical split/EFA/Procrustes/congruence pipeline on the permuted matrix ($20$ replicates $\times$ $6$ factors) yields mean $\phi{=}0.382$, $95$th percentile $0.668$, maximum $0.974$. A structureless matrix therefore returns $\phi \approx 0.38$, not $\approx 0.95$, and the observed values clear the null comfortably. Two caveats follow from having the distribution rather than the cutoff: the null has a long right tail, and the re-run's cross-corpus F4 ($0.9701$) sits below the null's single largest draw, so we do not describe any individual $0.97$ as ``excellent'' on the textbook band.

\paragraph{The Procrustes step is not doing the work.} Congruence is computed after a $6{\times}6$ orthogonal Procrustes alignment, which has $15$ free parameters and could in principle absorb real differences. Comparing the two half-sample Promax pattern matrices with no rotation at all gives $0.9880$--$0.9979$, against $0.9866$--$0.9987$ with the alignment: the free rotation buys $+0.001$. The two solutions are near-identical before it is applied.

\paragraph{A caveat on comparability.} The re-run is a fresh fit, not a re-scoring of the main text's solution, and its rotated six-factor solution is not label-identical to the one in Section~\ref{sec:r1}. Relative to the main text it merges the specificity and search-hardness axes into a single verifiable-constraint dimension, merges the density and length axes into a single asks-per-token dimension, and separates two axes (exemplar anchoring; heterogeneity of specificity within a prompt) that the main-text solution does not. The figures in Table~\ref{tab:rerun} are therefore internal-stability and cross-corpus statistics for the re-run's own solution: they establish that a stricter extractor yields a six-dimensional structure reproducing more cleanly, not that each main-text factor is individually unchanged. We report the discrepancy rather than harmonising it. It indicates that the F5/F6 instability hedged in Section~\ref{sec:r1} is a property of the extraction step rather than of articulation, and it means the identity of the axes at the margin is less settled than their number.

\section{Pre-registered Construct Validation of the Factor Labels}
\label{sec:appendix-construct}

The factor names in Section~\ref{sec:r1} are post-hoc descriptive labels, and a descriptive label can be wrong. This appendix tests them. The question is simple: if we deliberately edit a prompt in exactly the way a factor's name describes, does that factor's score move the way the name predicts?

\paragraph{Design and safeguards.} We wrote eight targeted manipulations of real prompts, one per construct: padding with verbal filler that adds no content, compressing the same content into far fewer words, replacing every vague requirement with a checkable one, hedging everything so the user sounds undecided, anchoring the ask to two known examples, naming three branded candidates, adding two sentences of personal background, and densifying with extra short requirements. Three safeguards keep the test honest. The predictions were written down and frozen before any edited prompt existed. The edited prompts ($180$ minimal pairs) were authored by a separate model shown only plain-language descriptions of the constructs --- never a loading, never the word ``factor''. And the edits were scored \emph{transform-only} through the already-fitted model, which we verified reproduces the pipeline's own scores at $|r|{=}1.0000$; nothing was refit. Each manipulation carries three frozen criteria --- the target factor moves in the predicted direction, no off-target factor moves further (diagonal dominance), and off-target movement stays within a stated bound (discriminant validity) --- giving $24$ criteria in total. Table~\ref{tab:construct-validation} reports the outcome.

\paragraph{The noise floor.} Section~\ref{sec:s1} describes Stage~1 as running at temperature $0$. That does not make it a function. Re-tagging the same $180$ prompts with the same model at the same temperature moves the factor scores by mean $|\Delta Z|$ of $0.029$ (self-disclosure) to $0.135$ (specificity heterogeneity), and flips $2$ of $180$ status decisions. Every effect below must be read against that floor, and we state it rather than assuming determinism.

\begin{table}[t]
\centering\footnotesize
\setlength{\tabcolsep}{3pt}
\begin{tabular}{llrrl}
\toprule
Manipulation & target & $\Delta$ & $d$ & verdict \\
\midrule
pad (filler only)      & density   & $-2.12$ & 1.55 & pass \\
compress               & density   & $+0.93$ & 1.08 & pass \\
specify (checkable)    & verif.\ spec. & $+1.09$ & 1.52 & pass \\
exemplify              & exemplar  & $-2.20$ & 4.59 & pass \\
name (3 brands)        & named-alt & $+1.41$ & 1.38 & pass \\
disclose               & self-disc.\ & $+0.84$ & 1.01 & pass \\
\midrule
soften (hedge all)     & (F3)      & $-0.79$ & 0.93 & \textbf{falsified} \\
densify                & density   & $+0.22$ & 0.37 & \textbf{invalid} \\
\bottomrule
\end{tabular}
\caption{Pre-registered known-groups results. $\Delta$ is the mean shift in the target factor's standardised score; $d$ is Cohen's $d$. Twenty of the twenty-four frozen criteria hold. The six passing manipulations move their named factor at $d{=}1.0$ to $4.6$, move it further than any other factor, and clear the extractor's test--retest noise floor by $9\times$ to $49\times$. Factor numbering here is that of the re-extracted solution (Appendix~\ref{sec:appendix-rerun}); we give construct names rather than indices for that reason.}
\label{tab:construct-validation}
\end{table}

\paragraph{The two failures, and what they bought.} \emph{Densify} was our error rather than the instrument's. It added intent spans ($+2.05$) \emph{and} tokens ($+10.0$), and the density construct is a ratio; we pushed numerator and denominator in the same direction and measured nothing. We report it as an invalid test that licenses no conclusion. The pad and compress manipulations were pre-registered separately afterwards, as a second round, and validate the same construct bidirectionally --- which is a stronger result than the original test would have given.

\emph{Soften} is the informative failure. We predicted it would raise F3, which we had named for soft or unsettled constraints. It lowered F3 instead ($-0.79$), and the manipulation itself worked exactly as intended (crystallization $+1.53$, hard-constraint rate $-0.38$, mean specificity $-0.50$), so the prediction was wrong rather than the instrument. Diagnosing it: \emph{specify} and \emph{soften} are opposite manipulations and both lowered F3. A hardness axis cannot do that. What the two share is that each makes the prompt's specificity \emph{uniform}, and both cut the within-prompt variance of specificity --- with which F3 correlates more strongly ($r{=}+0.489$) than with anything else. F3 appears to measure the \emph{mixture} of precise and vague requirements inside a single prompt: a user who says ``under \$800'' and ``something nice'' in the same breath. We flag clearly that this replacement label is a post-hoc reading of correlations, \emph{not} a validated one: the manipulation that would test it convergently --- make half the requirements precise and leave half vague --- has not been written or run. Five of six labels survive a test they could have failed; the sixth did fail, and its replacement is not yet validated.

\paragraph{Separability of the two length-related constructs.} The \emph{disclose} manipulation also moved the density factor ($-0.78$), which we pre-registered: adding words without adding asks must lower asks-per-token. Which of the two moved further flips between tagging runs, and the gap ($0.06$) is smaller than the noise floor ($0.07$), so \emph{disclose} cannot settle separability and we do not claim it does. \emph{Pad} settles it: pure filler, with no requirement and no fact about the person, moves density by $-2.12$ and self-disclosure by only $-0.11$. Under a manipulation that touches only length, one factor moves $19\times$ further than the other.

\section{Nested-Model Decomposition of the Segment Block}
\label{sec:appendix-nested}

Section~\ref{sec:impl} reports that the segment dummies are jointly significant on every outcome, and notes that at this sample size joint significance is a weak statement. This appendix asks the sharper question: how much does knowing a prompt's articulation segment add \emph{over and above} topic, assistant model, source corpus, and length? We fit $M_0$ with those four controls only, then $M_2$ adding the segment dummies, and report in Table~\ref{tab:nested} the incremental $R^2$ together with a likelihood-ratio (binary outcomes) or $F$ (continuous outcomes) test of the added block. The models are fit on the re-extracted pipeline of Appendix~\ref{sec:appendix-rerun} ($n{=}13{,}576$), which is why the sample size differs from the main text.

\begin{table}[t]
\centering\footnotesize
\setlength{\tabcolsep}{4pt}
\begin{tabular}{lrl}
\toprule
Outcome & $\Delta R^2$ & segment block \\
\midrule
Recommendation specificity & $+0.0219$ & $F{=}43.2$, $p{\approx}5\mathrm{e}{-}44$ \\
Coverage (search)          & $+0.0138$ & $\chi^2{=}258$, $p{\approx}1\mathrm{e}{-}53$ \\
Gave direct answer         & $+0.0103$ & $\chi^2{=}191$, $p{\approx}2\mathrm{e}{-}39$ \\
Number of options          & $+0.0062$ & $F{=}17.7$, $p{\approx}2\mathrm{e}{-}17$ \\
Asked for clarification    & $+0.0037$ & $\chi^2{=}27.6$, $p{\approx}4\mathrm{e}{-}05$ \\
Response length            & $+0.0024$ & $F{=}9.2$, $p{\approx}1\mathrm{e}{-}08$ \\
\bottomrule
\end{tabular}
\caption{Incremental contribution of the segment block over $M_0$ (topic $+$ assistant model $+$ source corpus $+$ $\log$ prompt tokens). McFadden's pseudo-$R^2$ for binary outcomes, ordinary $R^2$ for continuous. The increments are modest in absolute terms, which is what one should expect from a single categorical predictor added on top of strong controls, but they are systematic and jointly significant on every outcome: articulation type predicts response characteristics over and above what the prompt is about, who answered it, where it came from, and how long it was.}
\label{tab:nested}
\end{table}

\section{Computational Cost and Software}
\label{sec:appendix-cost}

\paragraph{Reproduction.} The repository linked in Section~\ref{sec:intro} contains the complete runnable Stage~0--Stage~7 pipeline with a documented quickstart, a \texttt{download\_data.py} that auto-fetches every corpus slice from Hugging Face into the expected layout, all configs and hyperparameters, and a smoke-test config for cheap end-to-end verification. Reproduction starts at \texttt{python src/download\_data.py}. We release no new dataset because none exists: every input is a public corpus, which is a deliberate design choice rather than an omission --- the claim being tested is about naturalistic traffic, and a corpus we constructed ourselves could not test it.

\paragraph{API calls.} Stage~0 LLM classification (\texttt{gpt-4o-mini}) processed approximately $535{,}000$ candidate prompts at an estimated cost of \$85. Stage~1 (span tagging) and Stage~2 (response profiling), both using \texttt{claude-haiku-4-5}, processed $16{,}447$ positives each at an estimated combined cost of \$30--\$50. Total API spend was $<$\$150.

\paragraph{Local compute.} The e5-large-v2 embedding pass (for cascade candidate scoring on $1.32\text{M}$ non-domain prompts) ran on a single consumer GPU (RTX 5070 Ti) for approximately $60$ minutes. Factor analysis, GMM fitting, bootstrap ARI ($B{=}100$), factor-loading bootstrap CIs ($B{=}200$), and all downstream regressions ran on a single CPU in under $10$ minutes combined.

\paragraph{Packages.} Factor analysis used \texttt{factor\_analyzer} (Promax rotation, principal-axis extraction); GMM and cluster metrics used \texttt{scikit-learn} (\texttt{GaussianMixture}, \texttt{adjusted\_rand\_score}, \texttt{silhouette\_score}, $k$ selected by BIC); compositional ilr transforms followed \citet{egozcue2003isometric} implemented in \texttt{numpy}; embeddings used \texttt{intfloat/e5-large-v2} via \texttt{sentence-transformers}; regression statistics used \texttt{statsmodels} (OLS, logistic with L1 fallback); correlation tests used \texttt{scipy.stats}.

\paragraph{Hyperparameters.} LLM extraction calls used temperature $0$ and \texttt{max\_tokens}{=}$2000$ (Stage~1), $1500$ (Stage~2), $10$ (Stage~0). Factor extraction used $k{=}6$ selected by parallel analysis (Section~\ref{sec:s3}); GMM used $k{=}6$ selected by BIC over $k\in\{2,\ldots,6\}$ with $4$ random restarts (Section~\ref{sec:s4}); the e5 embedding threshold $\tau{=}0.80$ was calibrated on a held-out $1{,}000$-prompt sample.

\end{document}